\documentclass{article}

\usepackage{arxiv}

\usepackage[utf8]{inputenc} 
\usepackage[T1]{fontenc}    

\usepackage{hyperref}       
\usepackage{url}            
\usepackage{booktabs}       
\usepackage{amsfonts}       
\usepackage{nicefrac}       
\usepackage{microtype}      
\usepackage{lipsum}
\usepackage{mathtools}

\usepackage{graphicx}
\graphicspath{ {./images/} }
\usepackage{mathrsfs}

\title{Top2Vec: Distributed Representations of Topics}

\author{
  Dimo Angelov \\
  \texttt{dimo.angelov@gmail.com} \\
}

\begin{document}
\maketitle

\begin{abstract}
Topic modeling is used for discovering latent semantic structure, usually referred to as topics, in a large collection of documents. The most widely used methods are Latent Dirichlet Allocation and Probabilistic Latent Semantic Analysis. Despite their popularity they have several weaknesses. In order to achieve optimal results they often require the number of topics to be known, custom stop-word lists, stemming, and lemmatization. Additionally these methods rely on bag-of-words representation of documents which ignore the ordering and semantics of words. Distributed representations of documents and words have gained popularity due to their ability to capture semantics of words and documents. We present \texttt{top2vec}, which leverages joint document and word semantic embedding to find \emph{topic vectors}. This model does not require stop-word lists, stemming or lemmatization, and it automatically finds the number of topics. The resulting topic vectors are jointly embedded with the document and word vectors with distance between them representing semantic similarity. Our experiments demonstrate that \texttt{top2vec} finds topics which are significantly more informative and representative of the corpus trained on than probabilistic generative models.

\end{abstract}


\section{Introduction}

The ability to organize, search and summarize a large volume of text is a ubiquitous problem in natural language processing (NLP). Topic modeling is often used when a large collection of text cannot be reasonably read and sorted through by a person. Given a corpus comprised of many texts, referred to as documents, a topic model will discover the latent semantic structure, or topics, present in the documents. Topics can then be used to find high level summaries of a large collection of documents, search for documents of interest, and group similar documents together.

A topic is the theme, matter or subject of a text; it is thing being discussed. Topics are often thought of as discrete values, such as \emph{politics}, \emph{science}, and \emph{religion}. However, this is not to the case since any of these topics can be further subdivided into many other sub-topics. Additionally, a topic like \emph{politics} can overlap with other topics, such as the topic of \emph{health}, as they can both share the sub-topic of \emph{health care}. Any of these topics, their combinations or variations can be described by some unique set of weighted words. As such, we assume that topics are \emph{continuous}, as there are infinitely many combinations of weighted words which can be used to represent a topic. Additionally, we assume that each document has its own topic with a value in that continuum. In this view, the document's topic is the set of weighted words that are most informative of its unique topic, which can be a combination of the colloquial discrete topics. 

A useful topic model should find topics which represent a high-level summary of the information present in the documents. Each topic's set of words should represent information contained in the documents. For example, one can infer from a topic containing the words \emph{warming}, \emph{global}, \emph{temperature}, and \emph{environment}, that the topic is \emph{global warming}. We define topic modeling to be the process of finding topics, as weighted sets of words, that best represent the information of the documents. 

In the remainder of this section we discuss related work and introduce distributed representations of topics. In Section 2 we describe the \texttt{top2vec} model. Section 3 describes topic information gain and summarizes our experiments, and we conclude in Section 4. 


\subsection{Traditional Topic Modeling Methods}

The most widely used topic modeling method is Latent Dirichlet Allocation (LDA) \cite{LDA}. It is a generative probabilistic model which describes each document as a mixture of topics and each topic as a distribution of words. LDA generalizes Probabilistic Latent Semantic Analysis (PLSA) \cite{PLSA} by adding a Dirichlet prior distribution over document-topic and topic-word distributions. 


LDA and PLSA \emph{discretize} the continuous topic space into $t$ topics and model documents as mixtures of those $t$ topics. These models assume the number of topics $t$ to be known. The discretization of topics is \emph{necessary} to model the relationship between documents and words. This is one of the greatest weakness of these models, as the number of topics $t$ or the way to estimate it is rarely known, especially for very large or unfamiliar datasets \cite{syntactic_tm, lda_coherence}.

Each topic produced by these methods is a distribution of word probabilities. As such, the highest probability words in a topic are usually words such as \emph{the}, \emph{and}, \emph{it} and other common words in the language \cite{lda_coherence}. These common words, also called stop-words, often need to be filtered out in order to make topics interpretable, and extract the informative topic words. Finding the set of stop-words that must be removed is not a trivial problem since it is both language and corpus specific \cite{lda_common_words}; a topic model trained on text about \emph{dogs} will likely treat \emph{dog} as a stop-word since it is not very informative.

LDA and PLSA use bag-of-words (BOW) representations of documents as input which ignore word semantics. In BOW representation the words \emph{Canada} and \emph{Canadian} would be treated as different words, despite their semantic similarity. Stemming and lemmatization techniques aim to address this problem but often make topics harder to understand. Moreover, stemming and lemmatization do not recognize the similarity of words like \emph{big} and \emph{large}, which do not share a word stem.

The authors of the LDA paper explicitly state: "We refer to the latent multinomial variables in the LDA model as topics, so as to exploit text-oriented intuitions, but we make no epistemological claims regarding these latent variables beyond their utility in representing probability distributions on sets of words." \cite{LDA}. The objective of probabilistic generative models like LDA and PLSA is to find topics which can be used to recreate the original document word distributions with minimal error. However, a large proportion of all text contains uninformative words which may not be considered topical. These models do not differentiate between informative and uninformative words as their goal is to simply recreate the document word distributions. Therefore, the high probability words in topics they find do not necessarily correspond to what a user would intuitively think of as being topical.







\subsection{Distributed Representations of Words and Documents}

In neural networks, a \emph{distributed representation} means each concept learned by the network is represented by many neurons. Each neuron therefore participates in the representation of many concepts. When a neural network's weights are changed to incorporate new knowledge about a concept, the changes affect the knowledge associated with other concepts that are represented by similar patterns \cite{Hinton-dist}. Distributed representation has the advantage of leading to automatic generalization of the concepts learned. Distributed representations are often central to NLP machine learning techniques for learning vector representations of words and documents. 

Another key idea behind learning vector representations of words and documents is the \emph{distributional hypothesis}. The essence of the idea is captured by John Rupert Firth who famously said "You shall know a word by the company it keeps" \cite{JR}. This statement implies that words with similar meanings are used in similar contexts.


The continuous skip-gram and BOW models \cite{word2vec_1, word2vec_2} known as \texttt{word2vec}, introduced \emph{distributed word representations} that capture syntactic and semantic word relationships. The \texttt{word2vec} neural network learns word similarity by predicting which adjacent words should be present to a given context word in a sliding window over each document. The learning task of \texttt{word2vec} embraces the idea of distributional semantics, as it learns similar word vectors for words used in similar contexts. It also learns distributed representation of words, in the form of vectors, which facilitates generalization of word representation. The \texttt{word2vec} model generated word vectors produced state-of-the-art results on many linguistics tasks compared to traditional methods \cite{word2vec_1, word2vec_2, baroni, word_embedding_comparison}.


There has been interest in methods of finding distributed word vectors that do not rely on neural networks. It has been shown that the skip-gram version of \texttt{word2vec} is implicitly factorizing a word-context pointwise mutual information (PMI) matrix \cite{levy_implicit}, based on this finding the authors proposed \emph{Shifted Positive} PMI word-context representation of words. This has inspired other methods such as \texttt{GloVe} \cite{glove}, which learn context and word vectors by factorizing a global word-word co-occurrence matrix. Although \texttt{word2vec} implicitly factorizes a word-context PMI matrix, what it \emph{explicitly} does is maximize the dot product between word vectors for words which co-occur while minimizing dot product between words which do not co-occur. Additionally it uses a neural network which takes advantage of its learned distributed representation of words. This allows the model to learn about all words simultaneously from a single training step on a context word \cite{neural_lang}. The ability of \texttt{word2vec} word vectors to capture syntactic and semantic regularities of language that other methods try to recreate is a result of the former points, as is its ability to scale to large corpora \cite{word2vec_1, continuous_ling}. As shown in \cite{baroni}, quantitative comparisons between neural and non-neural word vectors show that neural learned vectors consistently perform better. Results from \cite{levy_implicit, levy_pmi} show that at best non-neural methods achieve results on certain tasks that are on-par with neural methods by replicating hyper-parameters of neural methods like \texttt{word2vec}. These methods, however, lack the ability to scale to large corpora.

With the goal of overcoming the weaknesses of BOW representations of documents, the \emph{distributed paragraph vector} was proposed with \texttt{doc2vec} \cite{doc2vec}. This model extends \texttt{word2vec} by adding a paragraph vector to the learning task of the neural network. In addition to the context window of words, a paragraph vector is also used to predict which adjacent words should be present. The paragraph vector acts as a memory of the topic of the document; it informs each context window of what information is missing \cite{doc2vec}. The \texttt{doc2vec} model can learn distributed representations of varying lengths of text, from sentences to documents. The \texttt{doc2vec} model outperforms BOW models and produces state-of-the-art results on many linguistics tasks compared to traditional methods \cite{doc2vec, doc2vec_eval}. The \texttt{doc2vec} model was followed by many works on general language models \cite{bert, electra, GPT}.

\subsection{Distributed Representations of Topics}



A \emph{semantic space} is a spatial representation in which distance represents semantic association \cite{griffiths}. A lot of attention has been given to semantic embedding of words. Specifically, distributed word vectors generated by models like \texttt{word2vec} which have been shown to capture syntactic and semantic regularities of language \cite{word2vec_1,distributed_words}. 

The \texttt{doc2vec} model is capable of learning document and word vectors that are jointly embedded in the same space. It has been observed that doing so, or using pre-trained word vectors, improves the quality of the learned document vectors \cite{doc2vec_eval}. These jointly embedded document and word vectors are learned such that document vectors are close to word vectors which are semantically similar. This property can be used for information retrieval as word vectors can be used to query for similar documents. It can also be used to find which words are most similar to a document, or most representative of a document. As mentioned in \cite{doc2vec}, the paragraph or document vector acts as a memory of the topic of the document. Thus the most similar word vectors to a document vector are likely the most representative of the document's topic. This joint document and word embedding is a \emph{semantic embedding}, since distance in the embedded space measures semantic similarity between the documents and words.

In contrast to traditional BOW topic modeling methods, the semantic embedding has the advantage of learning the semantic association between words and documents. We argue that the semantic space itself is a \emph{continuous representation of topics}, in which each point is a different topic best summarized by its nearest words. In the jointly embedded document and word semantic space, a dense area of documents can be interpreted as many documents that have a similar topic. We use this assumption to propose \texttt{top2vec}, a distributed topic vector which is calculated from dense areas of document vectors. The number of dense areas of documents found in the semantic space is assumed to be the number of prominent topics. The topic vectors are calculated as the centroids of each dense area of document vectors. A dense area is an area of very similar documents, and the centroid, or topic vector, can be thought of as the average document most representative of that area. We leverage the semantic embedding to find the words which are most representative of each topic vector by finding the closest word vectors to each topic vector. 

The \texttt{top2vec} model produces jointly embedded topic, document, and word vectors such that distance between them represents semantic similarity. Removing stop-words, lemmatization, stemming, and a priori knowledge of the number of topics are not required for \texttt{top2vec} to learn good topic vectors. This gives \texttt{top2vec} a major advantage over traditional methods. The topic vector can be used to find similar documents and words can be used to find similar topics. The same vector algebra demonstrated with \texttt{word2vec} \cite{word2vec_1, word2vec_2} can be used between the word, document and topic vectors. The topic vectors allow for topic sizes to be calculated based on each document vector's nearest topic vector. Additionally topic reduction can be performed on the topic vectors to hierarchically group similar topics and reduce the number of topics discovered. 

The greatest difference between \texttt{top2vec} and probabilistic generative models is how each models a topic. LDA and PLSA model topics as distributions of words, which are used to recreate the original document word distributions with minimal error. This often necessitates uninformative words which are not topical to have high probabilities in the topics since they make up a large proportion of all text. In contrast a \texttt{top2vec} topic vector in the semantic embedding represents a prominent topic shared among documents. The nearest words to a topic vector best describe the topic and its surrounding documents. This is due to the joint document and word embedding learning task, which is to predict which words are most indicative of a document, which necessitates documents, and therefore topic vectors, to be closest to their most informative words. Our results show that topics found by \texttt{top2vec} are significantly more informative and representative of the corpus trained on than those found by LDA and PLSA.






\section{Model Description}

\subsection{Create Semantic Embedding} \label{joint_embedding}

\begin{figure}[h]
\centering
\includegraphics[scale=0.4]{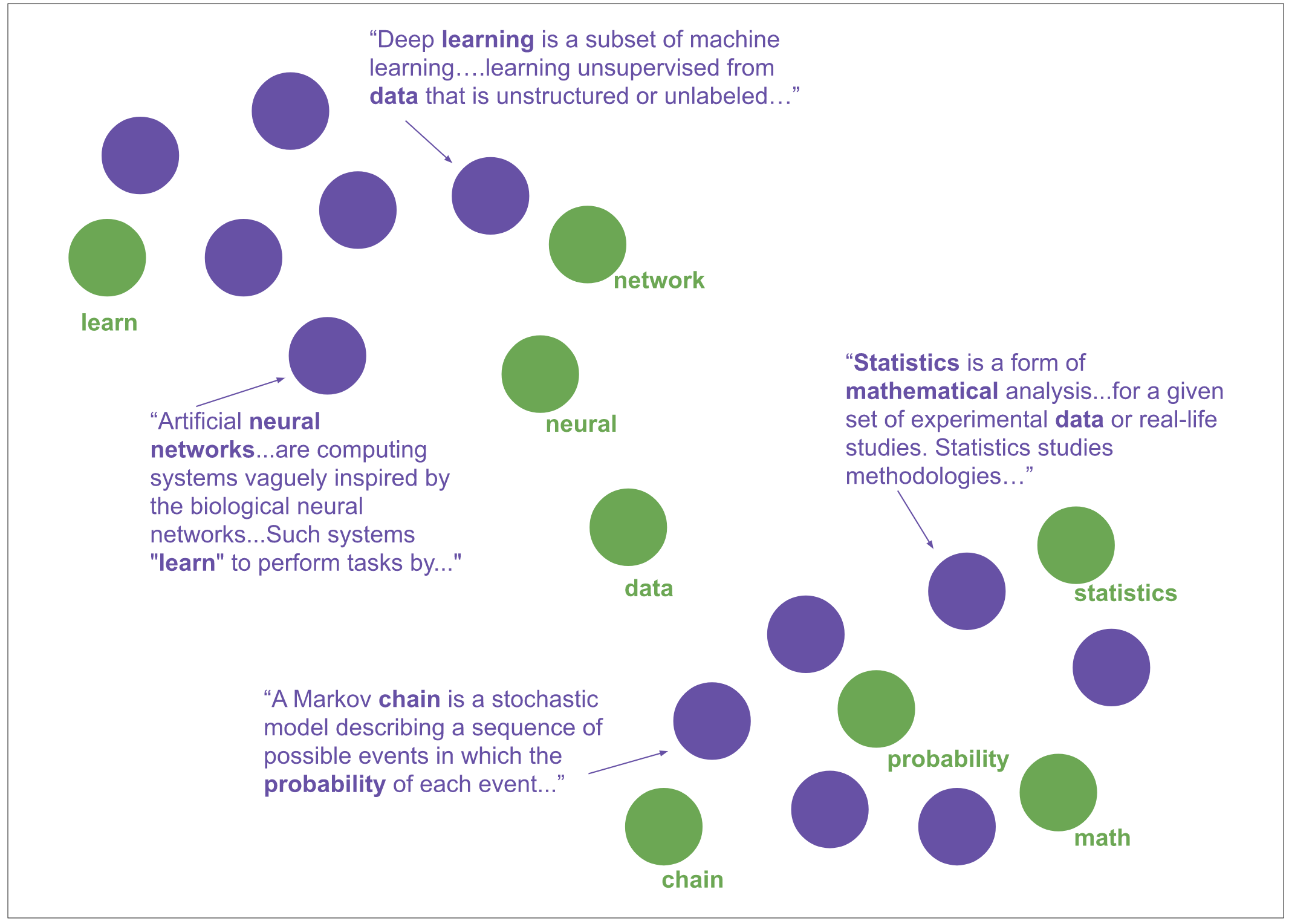}
\caption{An example of a semantic space. The purple points are documents and the green points are words. Words are closest to documents they best represent and similar documents are close together.}
\label{fig:semantic_space}
\end{figure}

In order to be able to extract topics, jointly embedded document and word vectors with certain properties are required. Specifically, we need an embedding where the distance between document vectors and word vectors represents semantic association. Semantically similar documents should be placed close together in the embedding space, and dissimilar documents should be placed further from each other. Additionally words should be close to documents which they best describe. With jointly embedded document and word vectors with these properties, topic vectors can be calculated. This spatial representation of words and documents is called a semantic space \cite{griffiths}. We argue that a semantic space with the outlined properties is a \emph{continuous representation of topics}. Figure \ref{fig:semantic_space} shows an example of a semantic space.

To learn jointly embedded document and word vectors we use \texttt{doc2vec} \cite{doc2vec, gensim}. There are two versions of the model: the Paragraph Vector with Distributed Memory (DM) and Distributed Bag of Words (DBOW). The DM model uses context words and a document vector to predict the target word within context window. The DBOW model uses the document vector to predict words within a context window in the document. Despite DBOW being a simpler model it has been shown to perform better \cite{doc2vec_eval}. Our experiments confirm these results and consequently we use the DBOW version of \texttt{doc2vec}.  

The \texttt{doc2vec} DBOW architecture is very similar to the \texttt{word2vec} skip-gram model which uses the context word to predict surrounding words in the context window. The only difference is that DBOW swaps the context word for the document vector, which is used to predict the surrounding words in the context window. This similarity allows for the training of the two to be interleaved, thus simultaneously learning document and word vectors which are jointly embedded. 




The key insight into how \texttt{doc2vec} and \texttt{word2vec} learn these vectors is understanding how the prediction task works specifically for DBOW and skip-gram models. The \texttt{word2vec} skip-gram model learns an input word and context word vector for each word in the vocabulary. The \texttt{word2vec} model consists of a matrix $W_{n,d}$ for input word vectors and $W'_{n,d}$ for context word vectors, where $n$ is the size of the corpus vocabulary, and $d$ is the size of the vectors to be learned for each word. Each row of $W_{n,d}$ contains a word vector $\vec{w} \in \mathbb{R}^d$ and each row of $W'_{n,d}$ contains a context word vector $\vec{w_c} \in \mathbb{R}^d$. For a given context window of size $k$, there will be $k$ words to the left and $k$ words to the right of the context word. For each of the $2k$ surrounding words $w$, their input word vector $\vec{w} \in W_{n,d}$  will be used to predict the context vector $\vec{w_c} \in W'_{n,d}$ of the context word $w_c$. For each surrounding word $w$ the prediction is $softmax(\vec{w} \cdot W_{n,d}')$. This generates a probability distribution over the vocabulary, for \emph{each} word being the context word $w_c$. The learning consists of using back propagation and stochastic gradient descent to update \emph{each} context word vector in $W_{n,d}'$, and $\vec{w}$ from $W_{n,d}$, such that the probability of the context vector given the surrounding word, $P(\vec{w_c}|\vec{w})$, is greatest in the probability distribution over the vocabulary. This process is repeated for every context window for all $n$ words. This learning process necessitates semantically similar words to have context word vectors which are close together while making dissimilar words have context word vectors which are distant. This is because in order to maximize the probability $P(\vec{w_c}|\vec{w})$, the value of $\vec{w}\cdot\vec{w_c}$ must be the maximum value in $\vec{w} \cdot W_{n,d}'$. This value is maximized when $\vec{w}$ is closest to $\vec{w_c}$ from word all context vectors in $W_{n,d}'$. Therefore the learning process updates $\vec{w}$ and $W_{n,d}'$ so that $\vec{w}$ and $\vec{w_c}$ are closer together. This can be interpreted as each context word pulling all similar context words towards it in the embedding space, while pushing away all dissimilar words. This results in a semantic space, represented by the context vectors $W'$, where all semantically similar words are close together and all dissimilar words are far apart.


The way the DBOW \texttt{doc2vec} model learns document vectors is similar to the \texttt{word2vec} skip-gram model. The model consists of a matrix $D_{c,d}$, where $c$ is the number of documents in the corpus and $d$ is the size of the vectors to be learned for each document. Each row of $D_{c,d}$ contains a document vector $\vec{d} \in \mathbb{R}^d$. The model also requires a context word matrix $W'_{n,d}$, which can be pre-trained, randomly initialized, or learned in parallel. For simplicity of the explanation, we will assume a scenario where matrix $W'_{n,d}$ has been pre-trained by a \texttt{word2vec} model on the same vocabulary of $n$ words. For each document $d$ in the corpus, the context vector $\vec{w_c} \in W'_{n,d}$ of each word in the document is used to predict the document's vector $\vec{d} \in D_{c,d}$. The prediction is $softmax(\vec{w_c} \cdot D_{c,d})$, which generates a probability distribution over the corpus for \emph{each} document being the document the word is from. The learning consists of using back propagation and stochastic gradient descent to update \emph{each} document vector in $D_{c,d}$ and $\vec{w_c}$ from $W_{n,d}'$, such that the probability of the document given the word, $P(\vec{d}|\vec{w})$, is greatest in the probability distribution over the corpus of documents. This learning process necessitates that document vectors be close to word vectors of words that occur in them and making them distant from word vectors of words that do not occur in them. This can be interpreted as each word attracting documents that are similar to them while repelling documents which are dissimilar to them. This results in a semantic space where documents are closest to the words that best describe them and far from words that are dissimilar to them. Similar documents will be close together in this space as they will be pulled into the same region by similar words. Dissimilar documents will be far apart as they will be attracted into different regions of the semantic space by different words.


We argue that the semantic space generated by \texttt{word2vec} and \texttt{doc2vec} is a \emph{continuous representation of topics}. This claim can be justified by observing what the learned vector space generated by \texttt{word2vec} represents. This model learns a matrix $W'_{n,d}$, which contains context word vectors of dimension $d$ for all $n$ words it is trained on. Each word vector in this matrix alone has no meaning; it only gives relative similarity to other word vectors in the matrix. We argue that this $d$ dimensional embedding space is a continuous representation of topics defined by the matrix $W'_{n,d}$. The matrix $W'_{n,d}$ can be seen as a linear transformation that when applied to a $d$ dimensional vector from the embedding space generates an $n$ dimensional vector. This vector itself is some measure of the strengths of each of the $n$ words in the vocabulary corresponding to the point in the $d$ dimensional space. However what this model has actually learned is how to transform points in the $d$ dimensional space into probability distributions over the $n$ words. Therefore any point $\vec{p}$, in the $d$ dimensional space can be transformed into a probability distribution over the $n$ word vocabulary using $softmax(\vec{p} \cdot W'_{n,d})$. Thus, any point in the $d$ dimensional space represents a different topic. Each word vector, $\vec{w_c} \in W'_{n,d}$, corresponds to the topic in the $d$ dimension space which has the greatest probability of word $w_c$. In general any point $\vec{p}$ in the $d$ dimensional space can be best described semantically by the nearest word vectors, since those are the words that would have the highest probability in its corresponding topic distribution over the $n$ words in the vocabulary.

There are several hyper-parameters that have a large impact on the performance of \texttt{doc2vec} \cite{doc2vec_eval}. The \emph{window size} is the number of words left and right of the context word. A \emph{window size} size of 15 has been found to produce the best results \cite{doc2vec_eval}, which our experiments support. The \texttt{doc2vec} model can use negative sampling or hierarchical softmax as its output layer. These are both meant to be efficient approximations of the full softmax \cite{word2vec_2}. We found that in our experiments the \emph{hierarchical softmax} produces better document vectors. According to \cite{doc2vec_eval}, the most important hyper-parameter is the \emph{sub-sampling threshold}, which determines the probability of high frequency words being discarded from a given context window. The suggested \emph{sub-sampling threshold} value is $10^5$. The smaller this number is, the more likely it is for a high frequency word to be discarded from the context window \cite{word2vec_2, doc2vec_eval}. A related hyper-parameter is \emph{minimum count}, which discards all words that have a total frequency that is less than that value from the model all together. This gets rid of extremely rare words which would not contribute to learning the document vectors. In our experiments we found a \emph{minimum count} of 50 to work best, however this value largely depends on corpus size and its vocabulary. The \emph{vector size} is the size of the document and word vectors that will be learned, the larger they are the more complex information they can encode. In general, the suggested \emph{vector size} is 300 \cite{doc2vec_eval}, with larger data sets larger values will lead to better results, at greater computational cost. The number of \emph{training epochs} suggested by \cite{doc2vec_eval} is 20 to 400, with the higher values for smaller data sets. We found 40 to 400 \emph{training epochs} to be a good range.

\subsection{Find Number of Topics}

The semantic embedding has the advantage of learning a continuous representation of topics. In the jointly embedded document and word vector space, with the properties outlined in \ref{joint_embedding}, documents and words are represented as positions in the semantic space. In this space each document vector can be seen as representing the topic of the document \cite{doc2vec}. The word vectors that are nearest to a document vector, are the most semantically descriptive of the document's topic.

In the semantic space, a dense area of documents can be interpreted as an area of highly similar documents. This dense area of documents is indicative of an underlying topic that is common to the documents. Since the document vectors represent the topics of the documents, the centroid or average of those vectors can be calculated. This centroid is the \emph{topic vector} which is most representative of the the dense area of documents it was calculated from. The words that are closest to this topic vector are the words that best describe it semantically. The main assumption behind \texttt{top2vec} is that the number of dense areas of document vectors equals the number of prominent topics. This is a natural way to discretize topics, since a topic is found for each group of documents sharing a prominent topic. 
 
In order find the dense areas of documents in the semantic space, density based clustering is used on the document vectors, specifically Hierarchical Density-Based Spatial Clustering of Applications with Noise (HDBSCAN) \cite{hdbscan, hdbscan_2, hdbscan_impl}. However, the "curse of dimensionality" which results from the high-dimensional document vectors introduces two main problems. In the high-dimensional semantic embedding space, regularly of 300 dimensions, the document vectors are very sparse. The document vector sparsity makes it difficult to find dense clusters and doing so comes at a high computational cost \cite{curse}. In order to alleviate these two problems, we perform dimension reduction on the document vectors with the algorithm Uniform Manifold Approximation and Projection for Dimension Reduction (UMAP) \cite{umap, umap_impl}. In the dimension-reduced space, HDBSCAN can then be used to find dense clusters of documents.

\subsubsection{Low Dimensional Document Embedding}

\begin{figure}[h]
\centering
\includegraphics[scale=0.45]{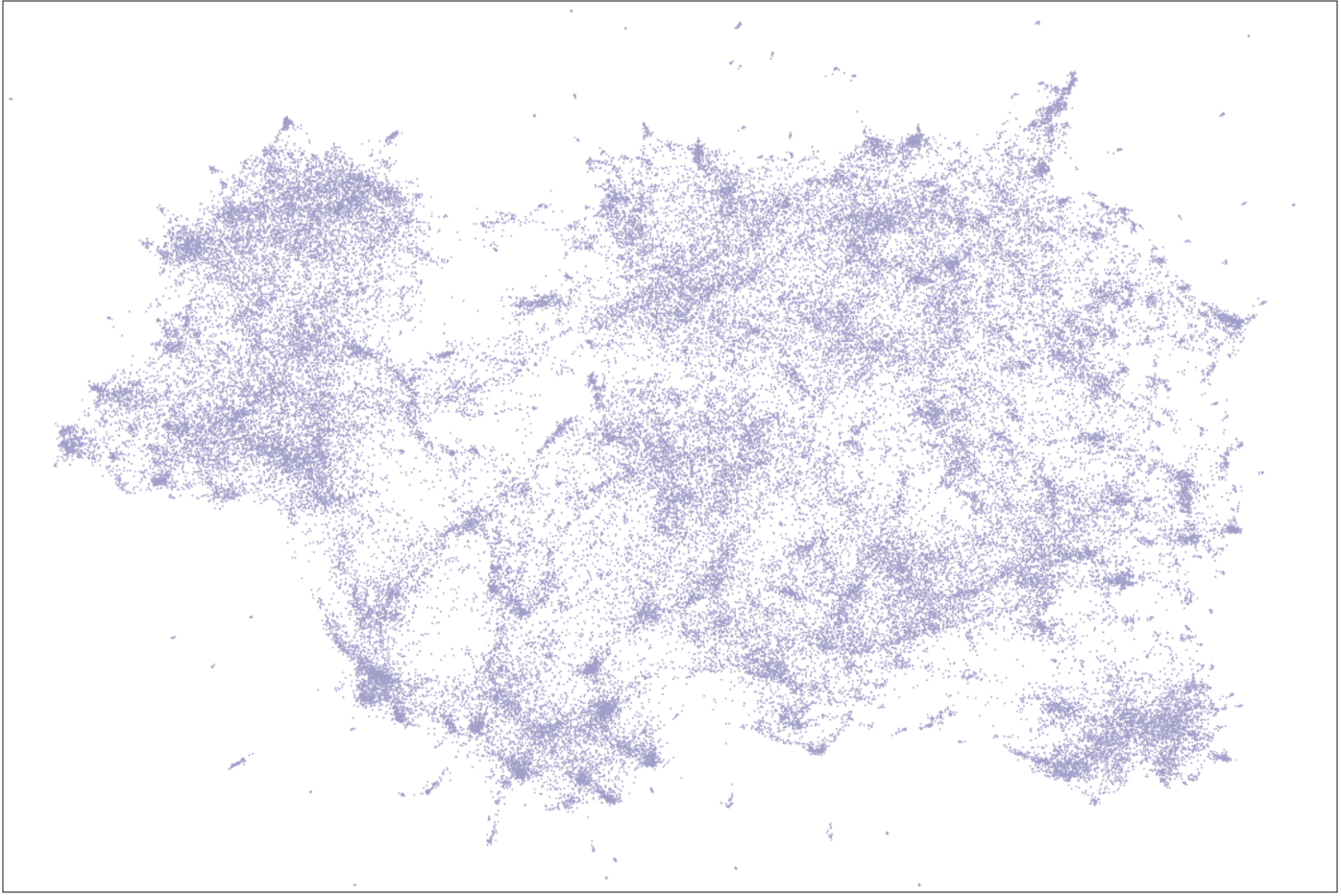}
\caption{300 dimensional document vectors from the \emph{20 news groups} dataset that are embedded into 2 dimensions using UMAP.}
\label{fig:low_d_embedding}
\end{figure}

Dimension reduction allows for dense clusters of documents to be found more efficiently and accurately in the reduced space. UMAP is a manifold learning technique for dimension reduction with strong theoretical foundations \cite{umap, umap_impl}. T-distributed Stochastic Neighbor Embedding (t-SNE) \cite{tsne} is another popular dimensional reduction technique. We found that t-SNE does not preserve global structure as well as UMAP and it does not scale well to large datasets. Hence, UMAP is chosen for dimension reduction in \texttt{top2vec}, as it preserves local and global structure, and is able to scale to very large datasets. Figure \ref{fig:low_d_embedding} shows UMAP-reduced document vectors; it can be seen that a lot of global and local structure is preserved in the embedding.

UMAP has several hyper-parameters that determine how it performs dimension reduction. Perhaps the most important parameter is the \emph{number of nearest neighbours}, which controls the balance between preserving global structure versus local structure in the low dimensional embedding. Larger values put more emphasis on global over local structure preservation. Since the goal is to find dense areas of documents which would be close to each other in the high dimensional space, local structure is more important in this application. We find that setting \emph{number of nearest neighbours} to 15 gives the best results, as this value gives more emphasis on local structure. Another related parameter is the \emph{distance metric}, which is used to measure the distance between points in the high dimensional space. The often used \emph{distance metric} for the document vectors is \emph{cosine similarity} \cite{word2vec_1, word2vec_2}, because it measures similarity of documents irrespective of their size. Lastly the \emph{embedding dimension} must be chosen; we find 5 dimensions to give the best results for the downstream task of density based clustering. 

\subsubsection{Find Dense Clusters of Documents}

\begin{figure}[h]
\centering
\includegraphics[scale=0.45]{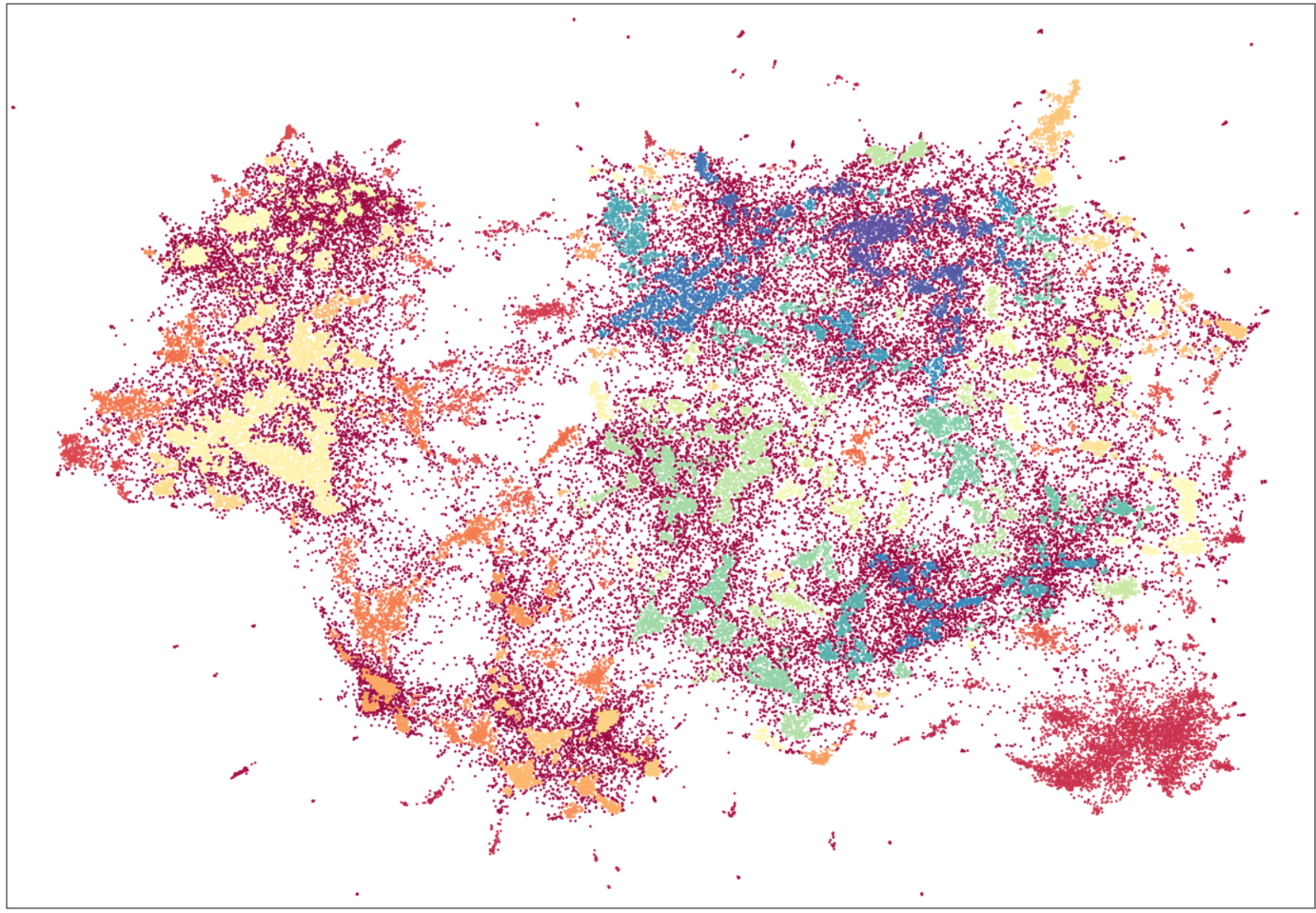}
\caption{UMAP-reduced document vectors from the \emph{20 news groups} dataset. Each colored area of points is a dense area of documents identified by HDBSCAN, red points are documents HDBSCAN has labeled as noise.}
\label{fig:dense_areas}
\end{figure}

The goal of density based clustering is to find areas of highly similar documents in the semantic space, which indicate an underlying topic. This is performed on the UMAP reduced document vectors. The challenge is that the document vectors will have varying density throughout the semantic space. Additionally there will be sparse areas where documents are highly dissimilar. This can be seen as noise, as there is no prominent underlying topic. In order to overcome these challenges, HDBSCAN is used to find the dense areas of document vectors, as it was designed to handle both noise and variable density clusters \cite{hdbscan_2}. HDBSCAN assigns a label to each dense cluster of document vectors and assigns a noise label to all document vectors that are not in a dense cluster. The dense areas of identified document vectors will be used to calculate the topic vectors. Documents that are classified as noise can be seen as not being descriptive of a prominent topic. Figure \ref{fig:dense_areas} shows an example of dense areas of documents identified by HDBSCAN.

The main hyper-parameter that needs be chosen for HDBSCAN is \emph{minimum cluster size}; this parameter is at the core of how the algorithm finds clusters of varying density \cite{hdbscan_2}. This parameter represents the smallest size that should be considered a cluster by the algorithm. We find that a \emph{minimum cluster size} of 15 gives the best results in our experiments, as larger values have a higher chance of merging unrelated document clusters.

\subsection{Calculate Topic Vectors}


\subsubsection{Calculate Centroids in Original Dimensional Space}

The dense clusters of documents and noise documents identified by HDBSCAN in the UMAP-reduced dimension, correspond to locations in the original semantic embedding space. The use of UMAP and HDBSCAN can be seen as a process which labels each document in the semantic embedding space with either a noise label or a label for the dense cluster to which it belongs.

\begin{figure}[h]
\centering
\includegraphics[scale=0.3]{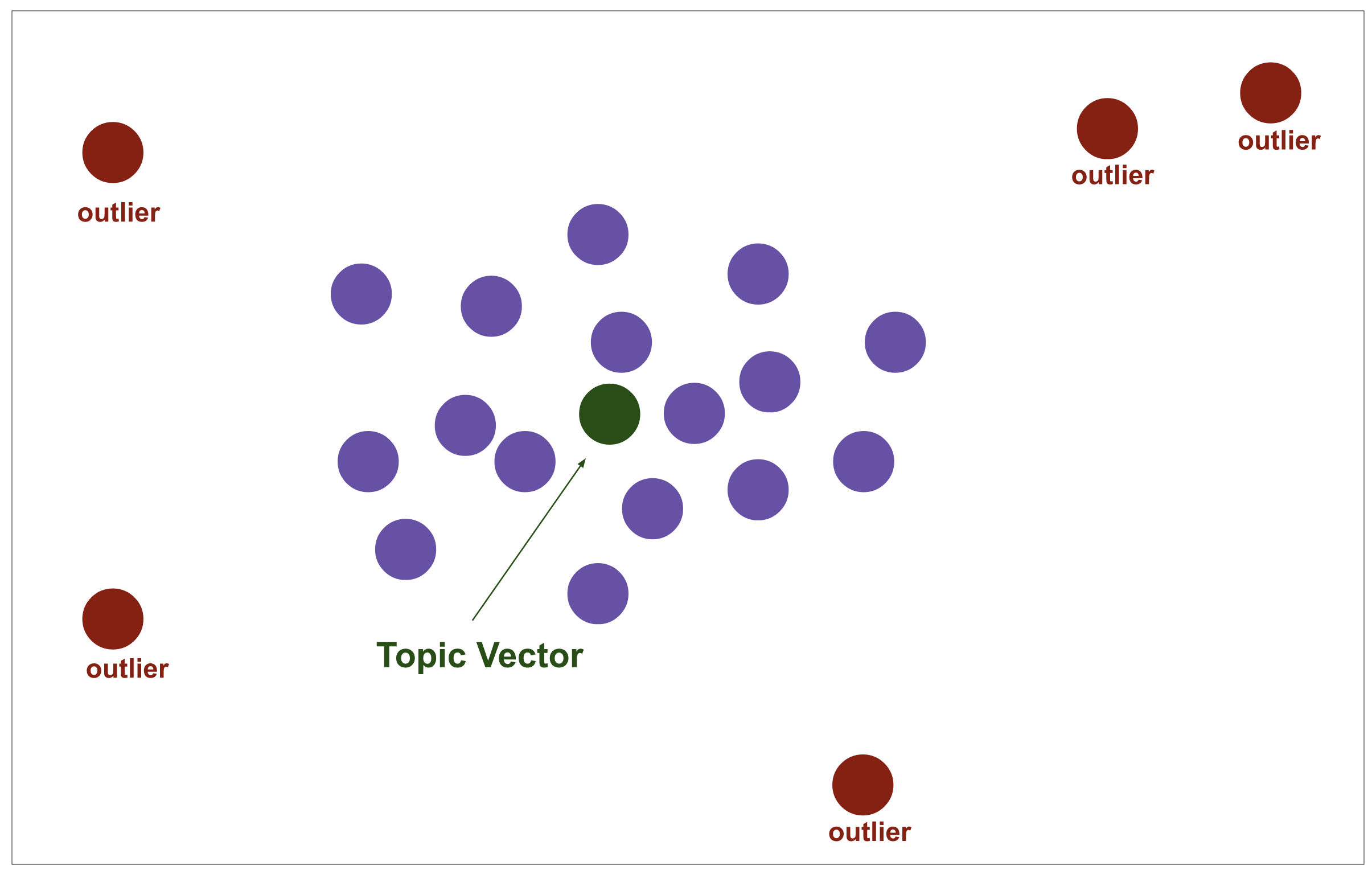}
\caption{The topic vector is the centroid of the dense are of documents identified by HDBSCAN, which are the purple points. The outliers identified by HDBSCAN are not used to calculate the centroid. }
\label{fig:topic_vector}
\end{figure}

Given labels for each cluster of dense documents in the semantic embedding space, topic vectors can be calculated. There are a number of ways that the topic vector can be calculated from the document vectors. The simplest method is to calculate the centroid, i.e. the arithmetic mean of all the document vectors in the same dense cluster. There are other reasonable options such as the geometric mean or using probabilities from the confidence of clusters created by HDBSCAN. We experimented with these techniques and found that they resulted in very similar topic vectors, with almost identical nearest-neighbour word vectors. We speculate that this is mainly due to the sparsity of the high dimensional space. Therefore, we decided to use the simple method of calculating the centroid. Figure \ref{fig:topic_vector} shows a visual example of a topic vector being calculated from a dense area of documents. 

The centroid is calculated for each set of document vectors that belong do a dense cluster, generating a topic vector for each set. The number of dense areas found is the number of prominent topics identified in the corpus.

\subsubsection{Find Topic Words}

\begin{figure}[h]
\centering
\includegraphics[scale=0.3]{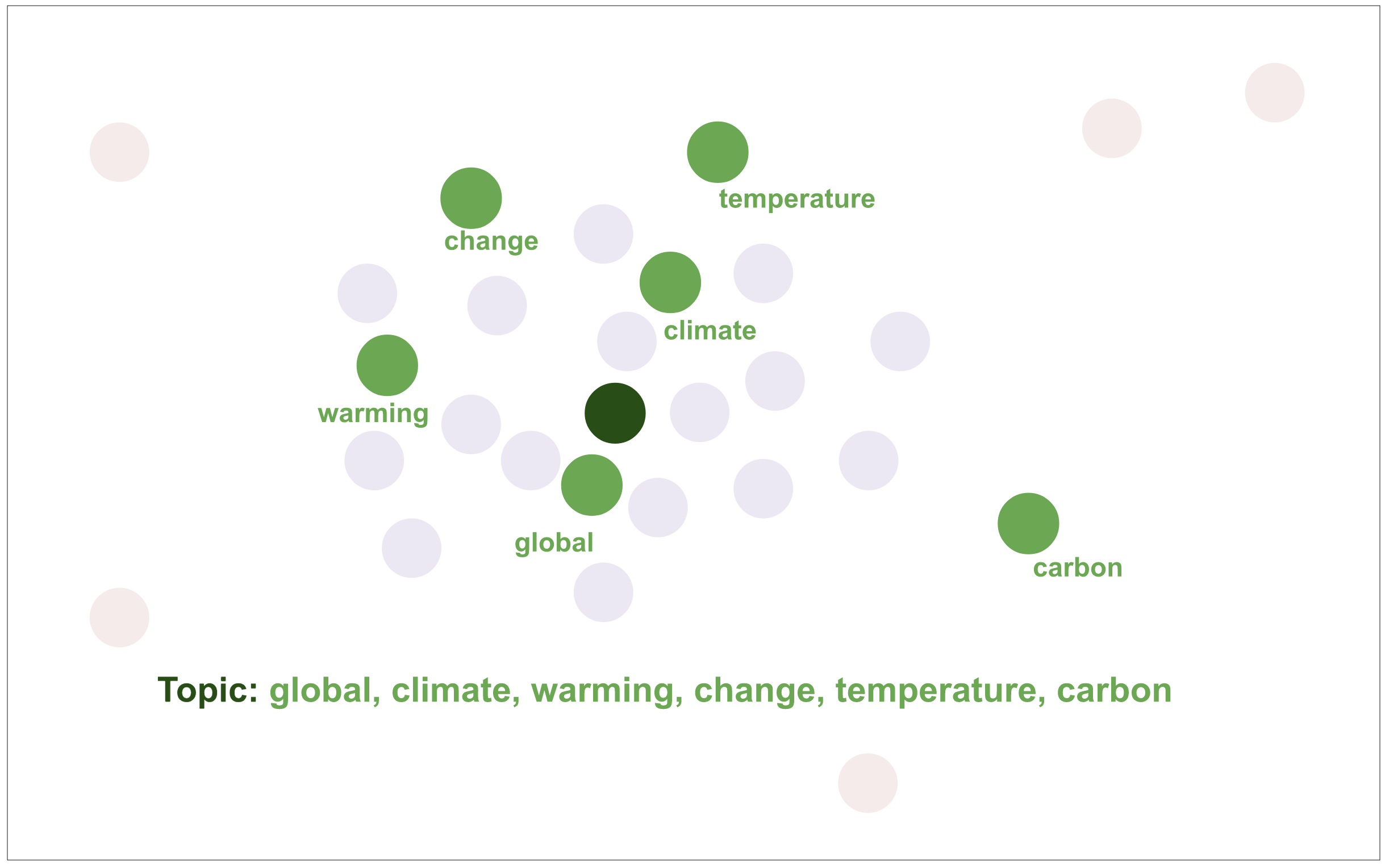}
\caption{The topic words are the nearest word vectors to the topic vector.}
\label{fig:topic_words}
\end{figure}

In the semantic space, every point represents a topic that is best described semantically by its nearest word vectors. Therefore the word vectors that are closest to a topic vector are those that are most representative of it semantically. The distance of each word vector to the topic vector will indicate how semantically similar the word is to the topic. The words closest to the topic vector can be seen as the words that are most similar to all documents into the dense area, as the topic vector is the centroid of that area. These words can be used to summarize the common topic of the documents in the dense area. Figure \ref{fig:topic_words} shows an example of a topic vector and the nearest words.

Common words appear in most documents and, as such, they are often in a region of the semantic space that is equally distant from all documents. As a result the words closest to a topic vector will rarely be stop-words, which has been confirmed in our experiments. Therefore there is no need for stop-word removal. 

\subsection{Topic Size and Hierarchical Topic Reduction}



The topic and document vectors allow for the size of topics to be calculated. The topic vectors can be used to partition the document vectors such that each document vector belongs to its nearest topic vector. This associates each document with exactly one topic, the one which is most semantically similar to the document. The size of each topic is measured as the number of documents that belong to it.

An advantage of the topic vectors and the continuous representation of topics in the semantic space is that the number of topics found by \texttt{top2vec} can be hierarchically reduced to any number of topics less than the number initially found. This is done by iteratively merging the smallest topic into its most semantically similar topic until the desired number of topics are reached. This is done by taking a weighted arithmetic mean of the topic vector of the smallest topic and its nearest topic vector, each weighted by their topic size. After each merge, the topic sizes are recalculated for each topic. This hierarchical topic reduction has the advantage of finding the topics which are most representative of the corpus, as it biases topics with greater size. 

\section{Results}

\subsection{Topic Information Gain}

A natural way to evaluate topic models is to score how well the topics describe the documents. This evaluation measures how informative the topics are to a user. We propose using \emph{mutual information} \cite{infogain} to measure the information gained about the documents when described by their topic words.

Traditional topic modeling methods discretize topic space and describe documents as a mixture of those topics. In order to evaluate a set of these topics $T$ generated from documents $D$, the total information gained is calculated for each document when described with the proportions of topics given by the topic model. 

In contrast, \texttt{top2vec} learns a continuous representation of topics and places documents in that space corresponding to their topic. A topic vector found by \texttt{top2vec} represents the topic common to a group of documents, or the average of their individual topics. In order to evaluate a set of \texttt{top2vec} topics $T$ generated from documents $D$, the documents are partitioned into sub-sets, with each sub-set corresponding to document vectors with the same nearest topic vector. Thus each document is assigned to exactly one topic. To evaluate these topics, the total information gained is measured for each of the sub-set of documents when described by the words nearest to their topic vector. 

The total information gained, or mutual information, about all documents \emph{D} when described by all words \emph{W}, is given by:

\begin{equation} \label{eq_I}
\begin{split}
I(D,W)=\displaystyle\sum_{d \in D}\sum_{w \in W}P(d,w)log\left(\frac{P(d,w)}{P(d)P(w)}\right)
\end{split}
\end{equation}

The contribution of each co-occurrence between a document \emph{d} and word \emph{w} to the information gain calculation can be seen as the \emph{probability-weighted amount of information} (PWI) \emph{d} and \emph{w} contribute to the total amount of information \cite{aizawa}, given by:

\begin{equation} \label{eq_PWI}
\begin{split}
PWI(d,w)=P(d,w)log\left(\frac{P(d,w)}{P(d)P(w)}\right)
\end{split}
\end{equation}



Topics are distributions over the entire vocabulary \emph{W}. However, in order to evaluate their usefulness to a user, we evaluate them using the top \emph{n} words of the topic. For evaluation where each document is assigned to only one topic, each topic $\emph{t}\in\emph{T}$, will have a set of \emph{n} words $W_{t}\subset\emph{W}$ and documents $D_{t}\subset\emph{D}$. The information gained about all documents when described by their corresponding topic is given by:

\begin{equation} \label{eq_PWI_1}
\begin{split}
PWI(T) & = \displaystyle\sum_{t \in T}\sum_{d \in D_{t}}\sum_{w \in W_{t}}P(d,w)log\left(\frac{P(d,w)}{P(d)P(w)}\right)  \\
 & = \displaystyle\sum_{t \in T}\sum_{d \in D_{t}}\sum_{w \in W_{t}}P(d|w)P'(w)log\left(\frac{P(d,w)}{P(d)P(w)}\right)
\end{split}
\end{equation}

In equation \ref{eq_PWI_1}, $P(w)$ is the marginal probability of the word $w$ across all documents $D$. It is used to calculate the \emph{log} term, which is the \emph{pointwise mutual information} \cite{pointwise} between $w$ and $d$. $P'(w)$ is the probability of topic word w, which is used to calculate the \emph{expected mutual information} \cite{infogain}, or the information gained about document $d$ given topic word $w$. The quantity we are measuring is the information gained about each document given its corresponding topic words as a prior. Therefore $P'(w)$ is 1 and can be omitted \cite{aizawa}, which gives rise to:


\begin{equation} \label{eq_PWI_2}
\begin{split}
PWI(T) = \displaystyle\sum_{t \in T}\sum_{d \in D_{t}}\sum_{w \in W_{t}}P(d|w)log\left(\frac{P(d,w)}{P(d)P(w)}\right)
\end{split}
\end{equation}

Alternatively the equation can be generalized for the case that each document is represented by multiple topics. In this case we replace $P'(w)$ with $P(t)$, which is the proportion of words to be used to represent document $d$ by topic $t$:

\begin{equation} \label{eq_PWI_general}
\begin{split}
PWI(T) = \displaystyle\sum_{d \in D}\sum_{t \in T}\sum_{w \in W_{t}}P(d|w)P(t)log\left(\frac{P(d,w)}{P(d)P(w)}\right)
\end{split}
\end{equation}

Using Equations \ref{eq_PWI_2} and \ref{eq_PWI_general}, different sets of topics can be compared. A greater quantity of information gain indicates that the topics $t \in T$ are more informative of their corresponding documents. If topics contain words such as \emph{the}, \emph{and}, and \emph{it} or other intuitively uninformative words, they will receive lower information gain values. This is in large part due to the $P(d|w)$ term in the calculation, since the probability of any specific document given a very common word is very low. Therefore, the information gained is also low. Words that are mostly present in the subset of documents corresponding to the topic lead to higher information gain as they are informative of those documents. Additionally, low values of information gain will be obtained if the topic model assigns topics to the wrong documents. Topic information gain measures the quality of the words in the topic and their assignment to documents. Therefore, Equations \ref{eq_PWI_2} and \ref{eq_PWI_general} give values that correspond with what is intuitively more informative. We argue that due to its information theoretic derivation, topic information gain is a good measure for evaluating topic models. 

\subsection{LDA, PLSA and Top2Vec Topic Information Gain}

In order to evaluate LDA, PLSA and \texttt{top2vec} topics we train all models on the same documents $D$ and vocabulary $W$. Since \texttt{top2vec} automatically finds the number of topics, we compare LDA, PLSA and \texttt{top2vec} on increasing numbers of topics up to the amount discovered by \texttt{top2vec}. 

For each comparison between a set of LDA-generated topics, $T_{LDA}$, PLSA-generated topics, $T_{PLSA}$ and \texttt{top2vec}-generated topics, $T_{top2vec}$, we use the same number of top $n$ topic words and the same number of topics. Thus, for each comparison between $T_{LDA}$, $T_{PLSA}$, and $T_{top2vec}$, we ensure the following: 

\begin{itemize}
  \item $|T_{LDA}|=|T_{PLSA}| = |T_{top2vec}| =  number\: of\: topics$
  \item $|W_{t}|=n, \forall W_{t} \in T_{LDA}, T_{PLSA}, T_{top2vec} = top\: n\: topic\: words$
\end{itemize}

\subsubsection{20 News Groups Dataset}

The \emph{20 News Groups} dataset \cite{sklearn} contains 18,831 posts labelled with the news group they were posted in. We trained \texttt{top2vec}, LDA and PLSA models on this dataset using the same pre-processing steps. LDA and PLSA models were trained with 10 to 100 topics, with intervals of 10. Hierarchical topic reduction was used on the 103 topics discovered by \texttt{top2vec}.

\begin{figure}[h]
\centering
\includegraphics[scale=0.6]{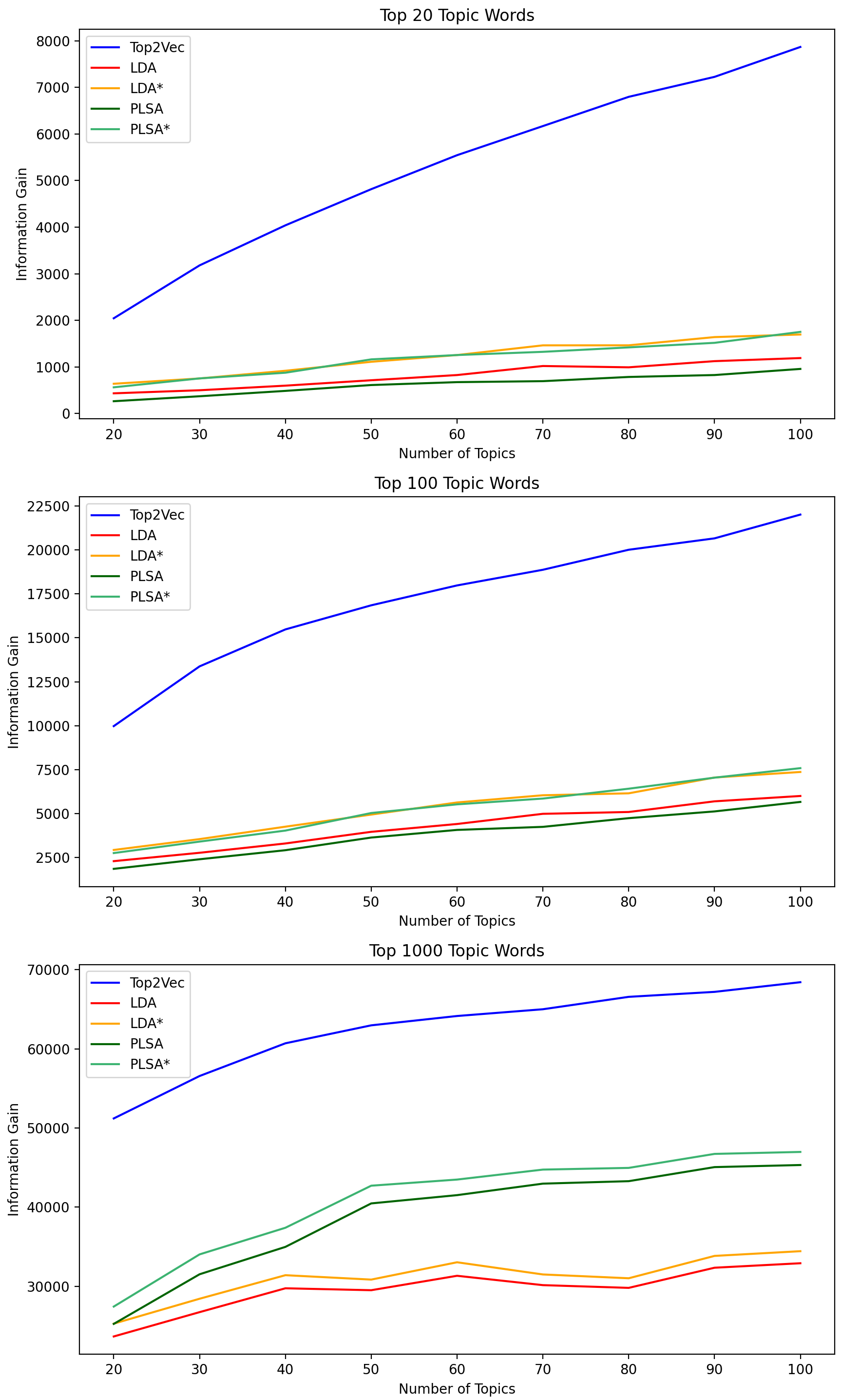}
\caption{Topic information gain comparison between Top2Vec, PLSA, and LDA trained models on the
\textit{20 News Groups} dataset. LDA* and PLSA* have stop-words removed.}
\label{fig:20_news_info_gain}
\end{figure}

To calculate $PWI(T_{LDA})$, $PWI(T_{PLSA})$, and $PWI(T_{top2vec})$, we use the same $W$ and $D$. A comparison of the topic information gain for models trained on the \textit{20 news groups} dataset can is shown in Figure \ref{fig:20_news_info_gain}. The results show that the top $n$ topic words from \texttt{top2vec} consistently provide more information than PLSA and LDA, with varying topic sizes and with up to the top 1000 topic words. Even when stop-words are filtered from LDA and PLSA. For most topic sizes the top 20 words from \texttt{top2vec} convey as much information as the top 100 from LDA and PLSA.


Tables \ref{tab:top2vec_topics} and \ref{tab:lda_topics} show the topics for \texttt{top2vec} and LDA models with topic size of 20. LDA was chosen over PLSA as it had higher topic information gain for 20 topics. Topics are ordered by increasing information gain. The topics shown for LDA have stop-words removed, where as the \texttt{top2vec} topics are the exact words discovered by the model. Tables \ref{tab:top2vec_topics} and \ref{tab:lda_topics} demonstrate that the topic information gain score corresponds to what is intuitively more informative. 

In Table \ref{tab:lda_topics}, LDA topics 2, 3 and 5 appear to contain nonsensical tokens, yet they have a high information gain. The \textit{20 news groups} data set contains messages that were sent encrypted or contain source code. When the \textit{20 news groups} messages are tokenized, these tokens are treated as words by the models. Thus, LDA has actually found informative tokens for that set of messages. However, that set contains only 23 messages. Therefore, LDA has found 3 different topics out of the requested 20, which only represent 23 messages out of the 18831 total amount the LDA model is trained on. This highlights an advantage of \texttt{top2vec} when finding the number of topics. 

\begin{figure}[h]
\centering
\includegraphics[scale=0.28]{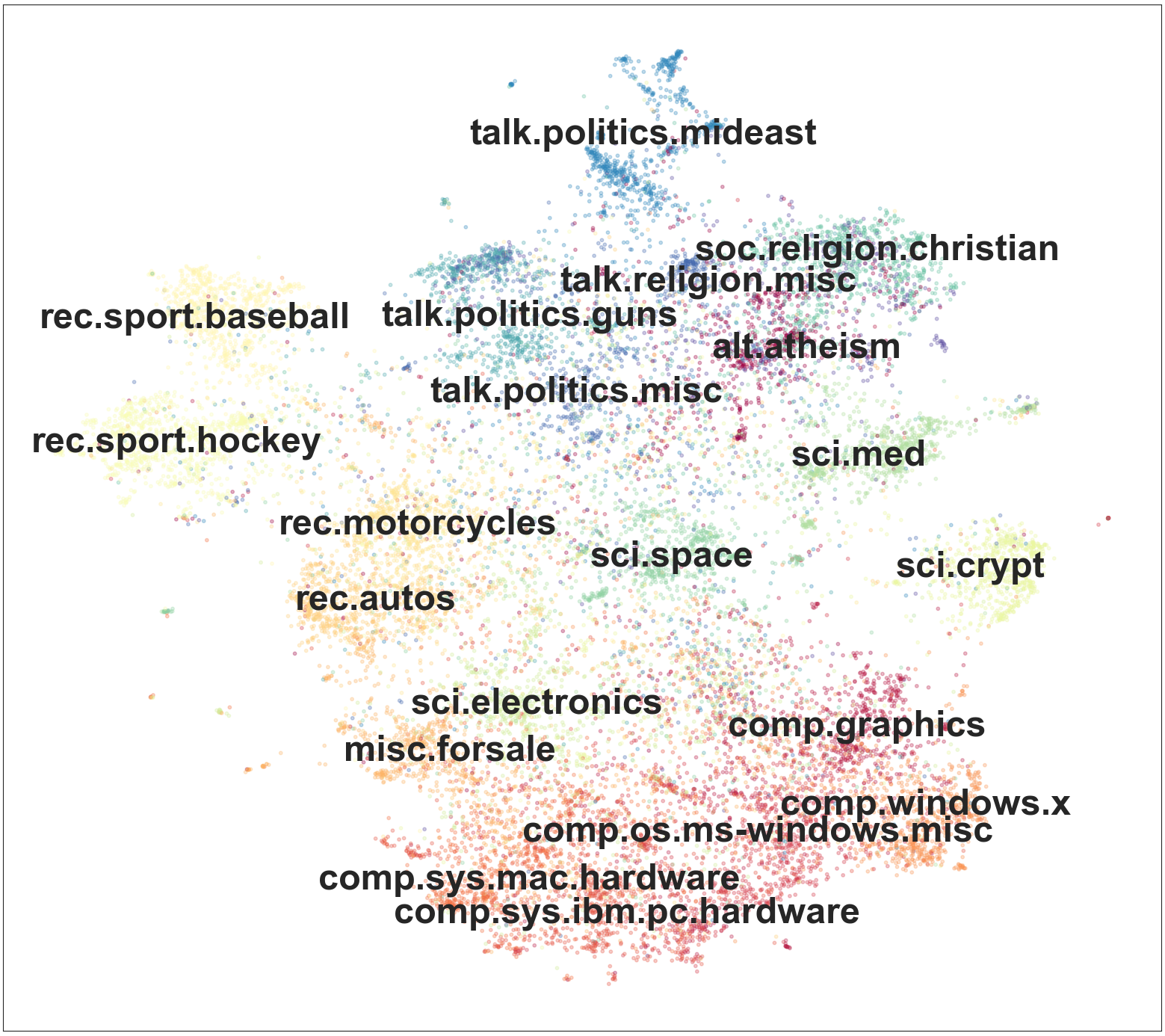}
\caption{Semantic embedding of \emph{20 news groups} messages labeled by news group. The 300 dimension document vectors are embedded into 2 dimensions using UMAP.}
\label{fig:20news_semantic}
\end{figure}

\begin{figure}[h]
\centering
\includegraphics[scale=0.28]{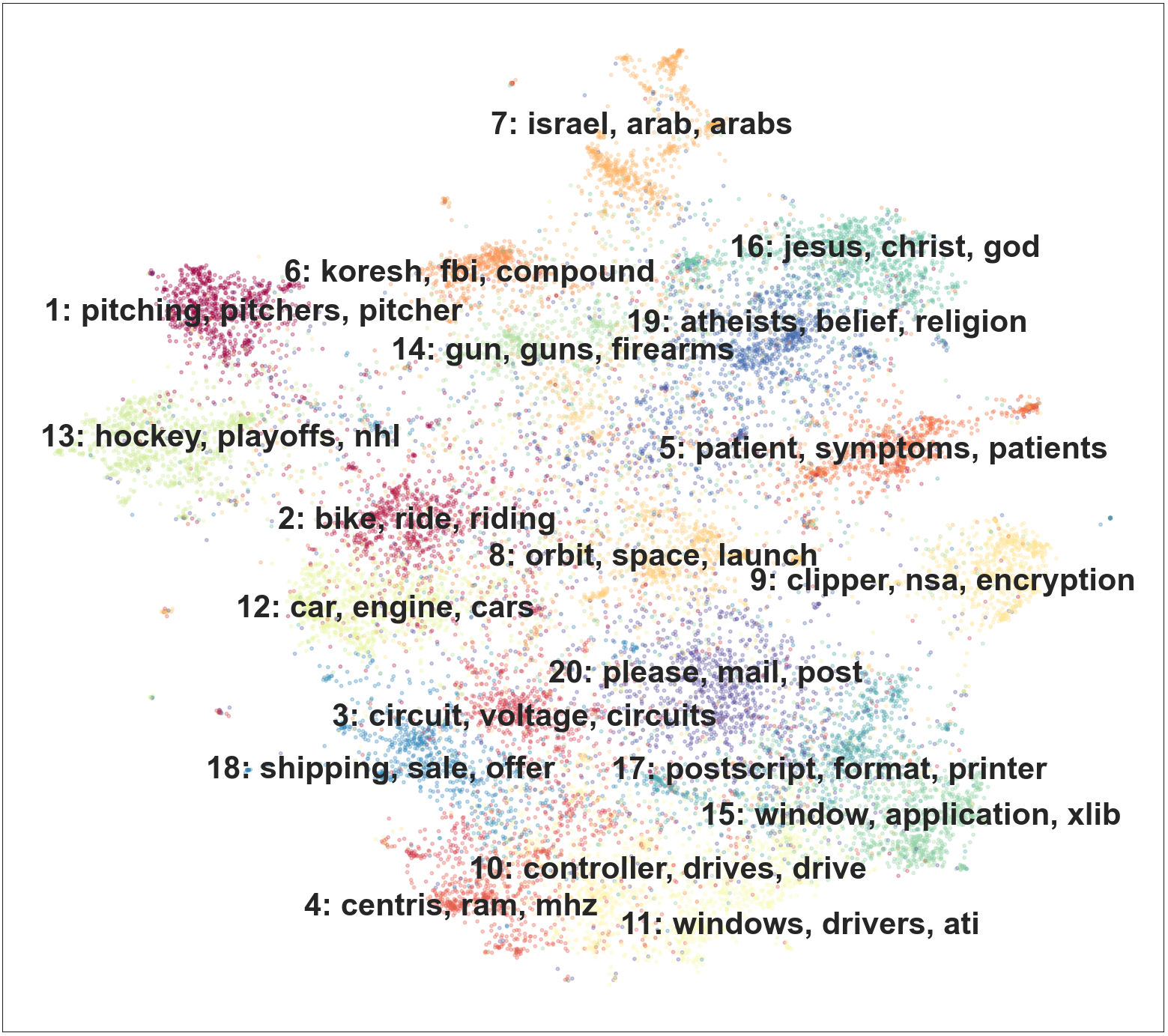}
\caption{The \emph{20 news groups} messages labeled with the \texttt{top2vec} topics from Table \ref{tab:top2vec_topics}. The 300 dimension document vectors are embedded into 2 dimensions using UMAP.}
\label{fig:20news_top2vec}
\end{figure}

Figure \ref{fig:20news_semantic} shows the semantic embedding of the messages labeled by the news group each message was posted in. This figure shows that the semantic embedding has learned the similarity of messages by visually demonstrating the continuous representation of topics. Messages from similar newsgroups are in similar regions of the semantic space. The small red points on the very right of Figure \ref{fig:20news_semantic}, are the 23 messages which predominantly contain encrypted content or large quantities of source code. Due to the density of that set of messages, \texttt{top2vec} finds a topic for those messages. However, when hierarchical topic reduction is performed to reduce the topic size to 20, due to its small size, the topic of the encrypted and source code containing messages is merged into another topic that is most semantically similar to it. The semantic embedding of the messages labeled with the 20 \texttt{top2vec} topics from Table \ref{tab:top2vec_topics} that each belong to is shown in Figure \ref{fig:20news_top2vec}. It demonstrates the assignment of the posts to the 20 topics correspond almost exactly to the 20 news groups and that each topic's top 3 words are very informative of the news group's actual topic. This visually demonstrates that \texttt{top2vec} finds topics which are more representative of the corpus as a whole, as confirmed by the topic information gain score in Figure \ref{fig:20_news_info_gain}. 

\begin{figure}[h]
\centering
\includegraphics[scale=0.4]{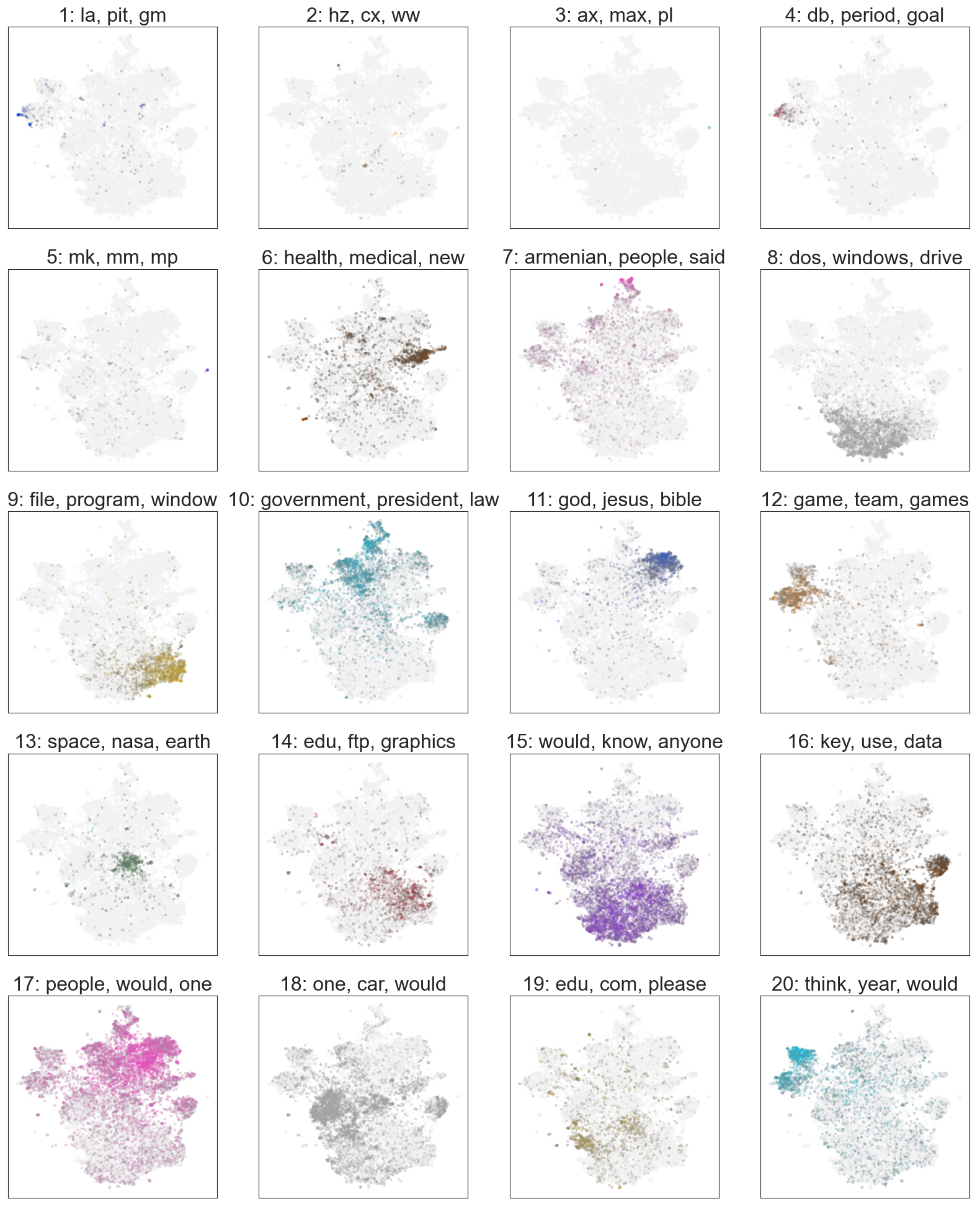}
\caption{Topic proportion of each LDA topic from Table \ref{tab:lda_topics} across all \emph{20 news groups} messages in the semantic embedding. The topics are ordered by decreasing information gain. The 300 dimension document vectors are embedded into 2 dimensions using UMAP.}
\label{fig:lda_topics_strengths}
\end{figure}

Topic information gain measures how informative topic words are of documents. Therefore, low scores are achieved when uninformative topic words are chosen, as well as when topics are assigned either to wrong documents or with incorrect proportions. There are a number of LDA topics in Table \ref{tab:lda_topics} that appear to be very coherent and that correspond to specific news groups. However, they have low scores in comparison to similar \texttt{top2vec} topics in Table \ref{tab:top2vec_topics}. This is explained, in part, by LDA's modeling of documents as a mixture of topics. It models each document with non-zero probabilities of all topics. Therefore each of the messages will have some non-zero proportion of the topics 2, 3 and 5 that were generated from encrypted or source code containing messages. Figure \ref{fig:lda_topics_strengths} shows the contribution of each LDA topic from Table \ref{tab:lda_topics} to all messages. It demonstrates that the most informative topics are highly localized and that the uninformative topics are spread out over many messages. Topic 15 and 17, which both have low information gain, make up a large proportion of most messages. These are topics with very generic words that are found in most documents.

The goal of LDA is to find topics such that their words recreate the original document word distributions with minimal error. This includes stop-words such as \emph{the}, \emph{and}, \emph{it} and other generic words that would not be considered informative or topical by a user. This explains topic 15 and 17 which are just the generic words that occur in most documents. LDA's goal can also result in extremely specific topics, such as 2, 3, and 5, which necessitate other topics to be more general. Figure \ref{fig:lda_topics_strengths} visually demonstrates the reason that LDA topics produce lower information gain; it finds many unlocalized and therefore uninformative topics compared to \texttt{top2vec}.



\begin{figure}[h]
\centering
\includegraphics[scale=0.25]{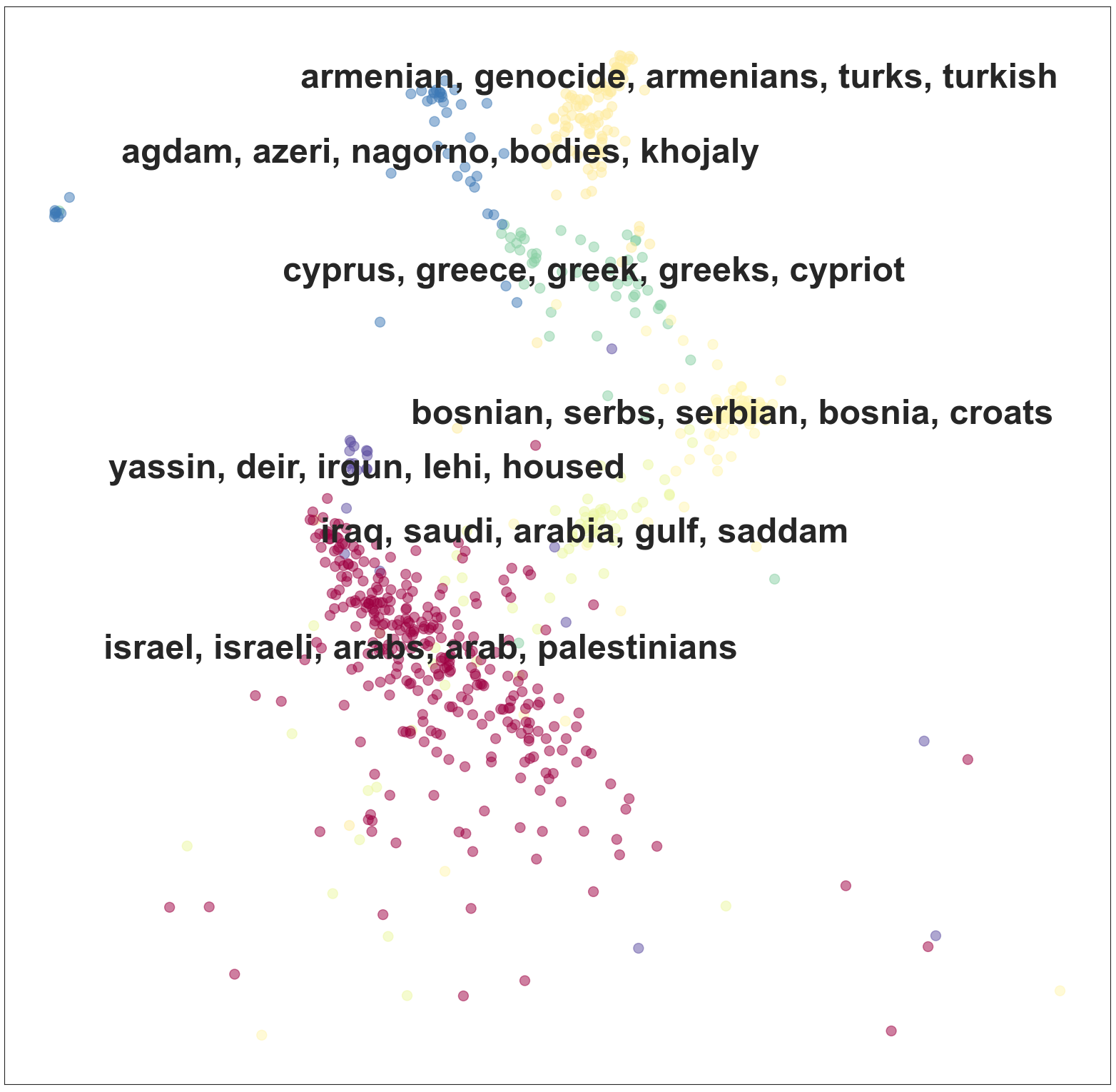}
\caption{Zoom in of \texttt{top2vec} original topics found in region of topic 7 from Table \ref{tab:top2vec_topics}. This region of the semantic space corresponds to the \emph{talk.politics.mideast} news group. The 300 dimension document vectors are embedded into 2 dimensions using UMAP.}
\label{fig:topic_7_zoom}
\end{figure}
\begin{figure}[h]
\centering
\includegraphics[scale=0.25]{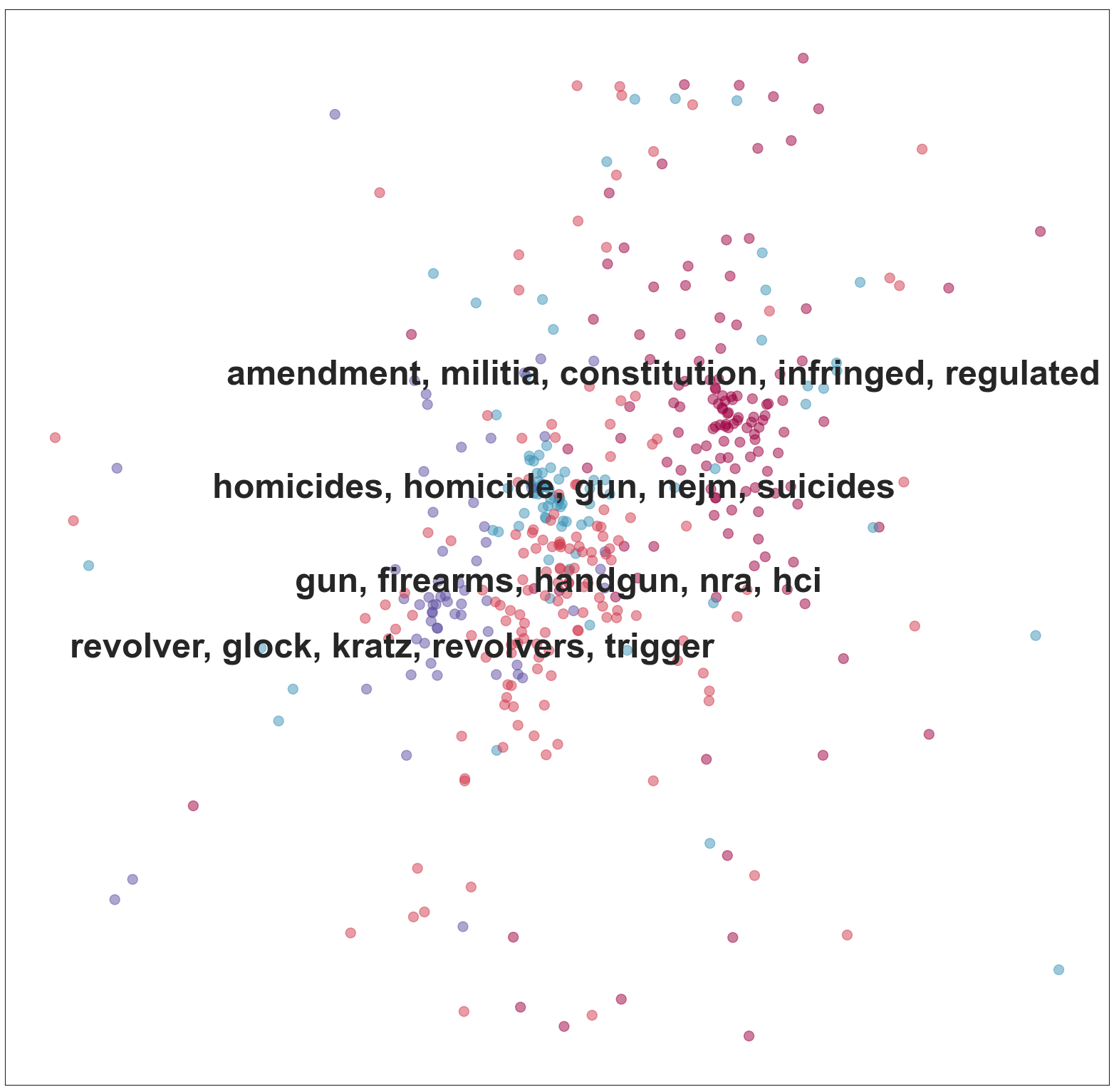}
\caption{Zoom in of \texttt{top2vec} original topics found in region of topic 14 from Table \ref{tab:top2vec_topics}. This region of the semantic space corresponds to the \emph{talk.politics.guns} news group. The 300 dimension document vectors are embedded into 2 dimensions using UMAP.}
\label{fig:topic_13_zoom}
\end{figure}

Figure \ref{fig:20_news_info_gain} shows that as the number of topics increases, the topic information gain for \texttt{top2vec} is consistently higher than for LDA and PLSA. This is because \texttt{top2vec} topics are more localized in the semantic space and therefore more informative. The number of topics found by \texttt{top2vec} on the \textit{20 News Groups} data set is 103, and are even more localized than the 20 topics in Table \ref{tab:top2vec_topics} which were generated from hierarchichal topic reduction. The original topics discovered in the region of topic 7 and 14 are shown in Figure \ref{fig:topic_7_zoom} and Figure \ref{fig:topic_13_zoom}. These topics are even more localized than the reduced topics and therefore more informative as indicated by the information gain scores in Figure \ref{fig:20_news_info_gain}.

\subsubsection{Yahoo Answers Dataset}

The \emph{Yahoo Answers} dataset \cite{YahooAnswers, YahooAnswers_2}, contains 1.3 million labelled posts. The posts are from 10 different topics, with 130,000 posts per topic. The number of topics \texttt{top2vec} found in this dataset are 2,618. Due to the computational cost of training LDA and PLSA models, we were only able to train the models from 10 to 100 topics with intervals of 10. Hierarchical topic reduction was used on the topics discovered by \texttt{top2vec}.

\begin{figure}[h]
\centering
\includegraphics[scale=0.6]{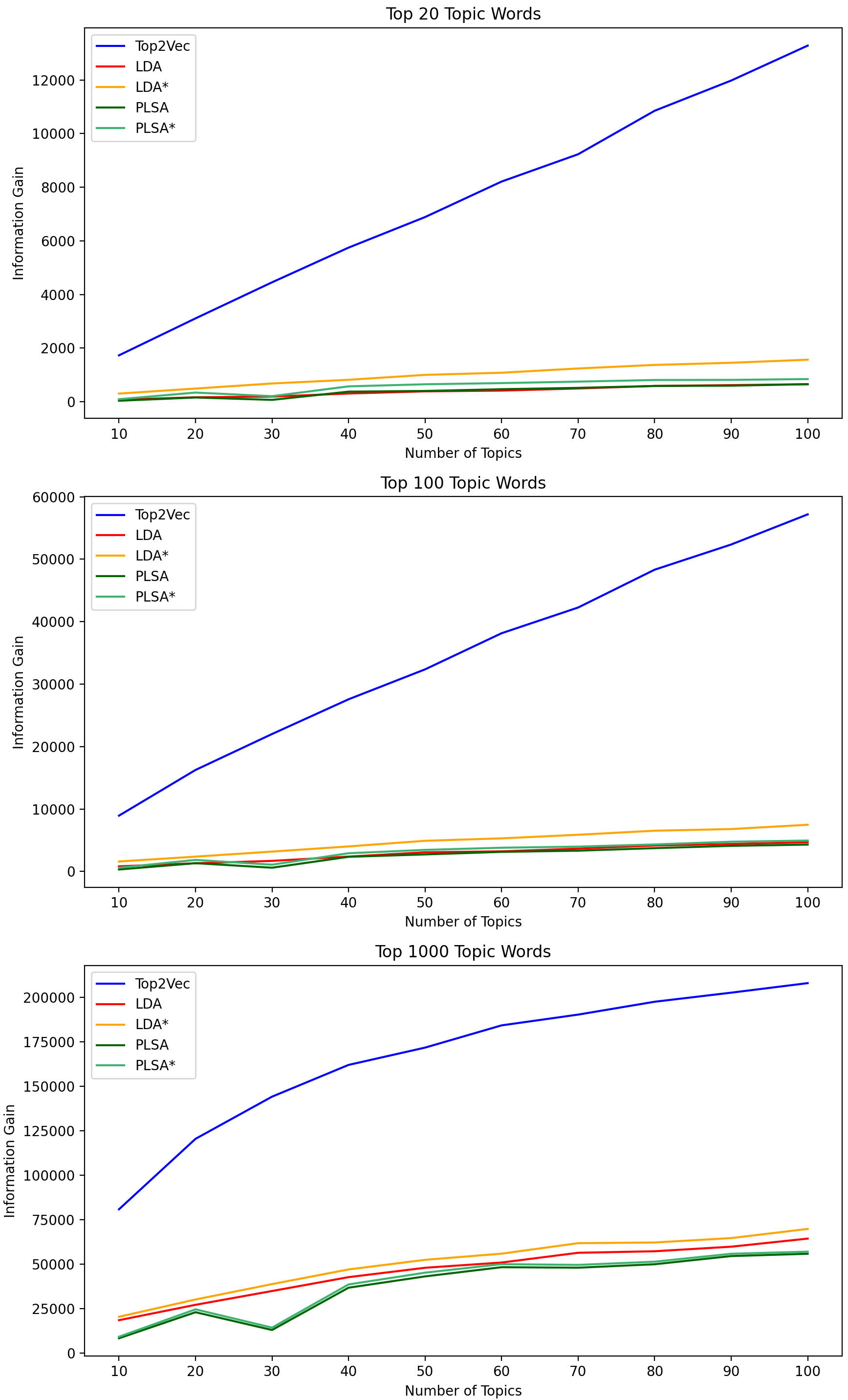}
\caption{Topic information gain comparison between Top2Vec, PLSA, and LDA trained models on the \textit{Yahoo Answers} dataset. LDA* and PLSA* have stop-words removed.}
\label{fig:yahoo_infogain}
\end{figure}

To calculate $PWI(T_{LDA})$, $PWI(T_{PLSA})$, and $PWI(T_{top2vec})$, we use the same $W$ and $D$. A comparison of the topic information gain for models trained on the \textit{Yahoo Answers} dataset can be seen in Figure \ref{fig:yahoo_infogain}. These results are consistent with the results from the \emph{20 News Groups} dataset. They show that the top $n$ topic words from \texttt{top2vec} consistently provide more information than PLSA and LDA, with varying topic sizes and up to the top 1000 topic words. Even when stop-words are filtered from LDA and PLSA. For most topic sizes, the top 20 words from \texttt{top2vec} convey as much information as the top 100 from LDA and PLSA. 

Tables \ref{tab:top2vec_topics_yahoo} and \ref{tab:lda_topics_yahoo} show the topics for \texttt{top2vec} and LDA models with a topic size of 10. The topics are ordered by increasing information gain. LDA was chosen over PLSA because it had higher topic information gain for 10 topics. The topics shown for LDA have stop-words removed, where as the \texttt{top2vec} topics are the exact words discovered by the model. This comparison demonstrates the interpretability of the topics and their associated information gain score, showing that the more informative topics receive higher information gain.

\begin{figure}[h]
\centering
\includegraphics[scale=0.28]{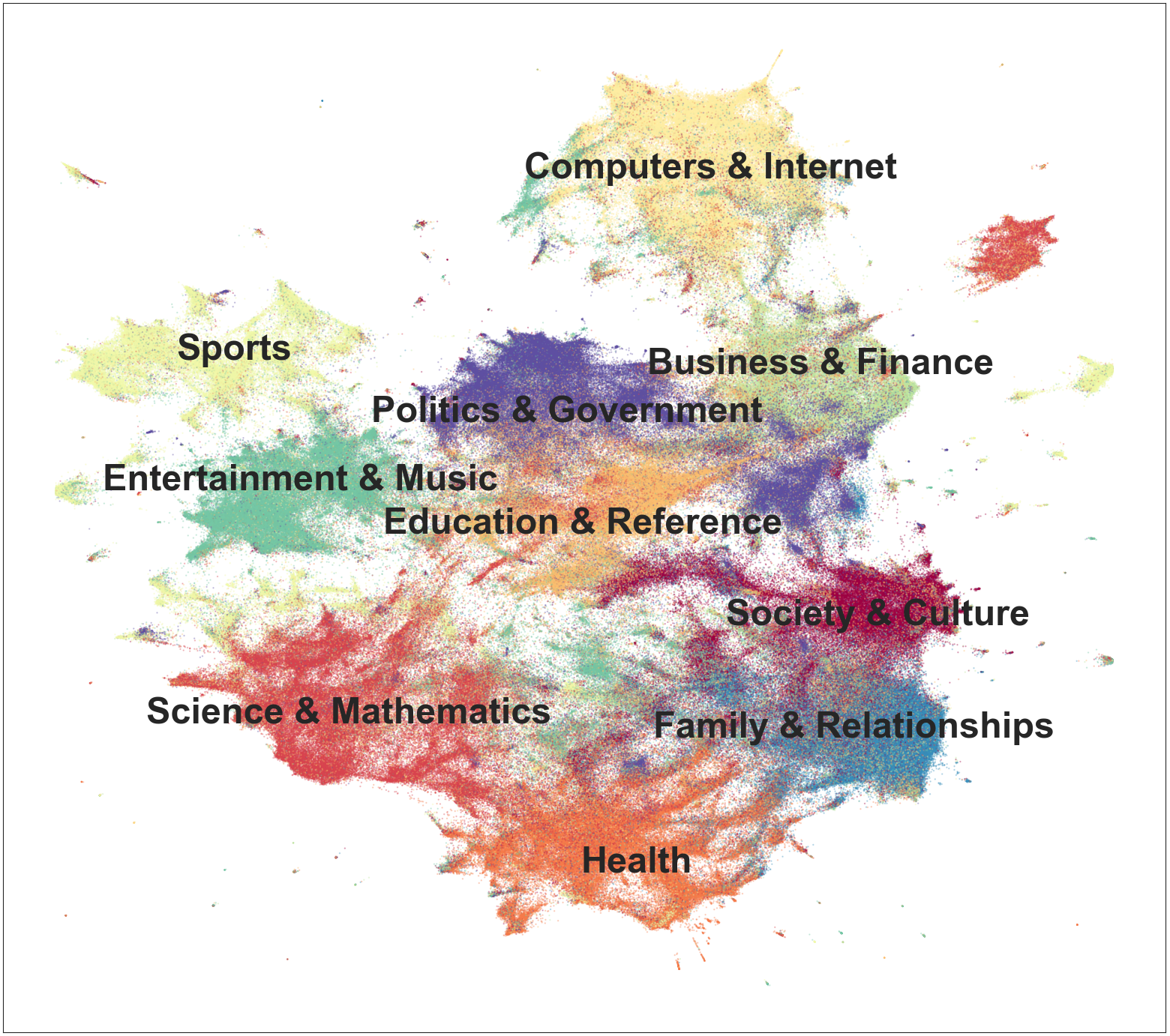}
\caption{Semantic embedding of \emph{Yahoo Answers} posts with true labels. The 300 dimension document vectors are embedded into 2 dimensions using UMAP.}
\label{fig:yahoo_semantic}
\end{figure}

\begin{figure}[h]
\centering
\includegraphics[scale=0.28]{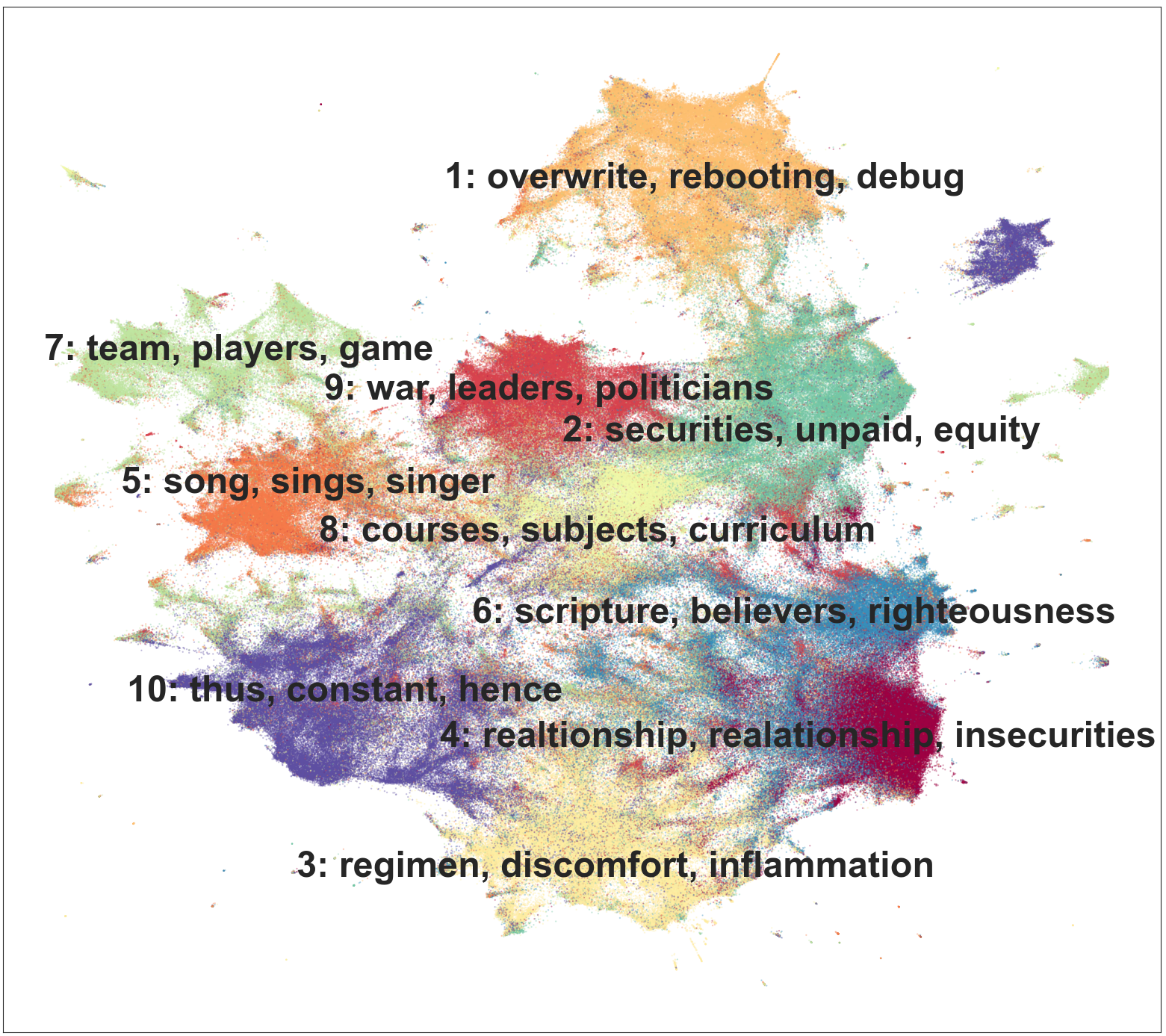}
\caption{\emph{Yahoo Answers} posts labeled with the \texttt{top2vec} topics from Table \ref{tab:top2vec_topics_yahoo}. The 300 dimension document vectors are embedded into 2 dimensions using UMAP.}
\label{fig:yahoo_top2vec}
\end{figure}

Figure \ref{fig:yahoo_semantic} shows the semantic embedding of \emph{Yahoo Answers} posts with their true topic labels. This figure demonstrates that the semantic space has captured the similarity of posts that share a similar topic. Figure \ref{fig:yahoo_top2vec} shows the posts labelled with the \texttt{top2vec} topics from Table \ref{tab:top2vec_topics_yahoo}. It demonstrates that the assignment of the posts to the 10 topics correspond almost exactly to the 10 topic labels from the \emph{Yahoo Answers} dataset and that the topic's top 3 words are very informative of the true topic. This visually demonstrates that \texttt{top2vec} finds topics that are representative of the corpus as a whole, as confirmed by the topic information gain score in Figure \ref{fig:yahoo_infogain}. 

\begin{figure}[h]
\centering
\includegraphics[scale=0.43]{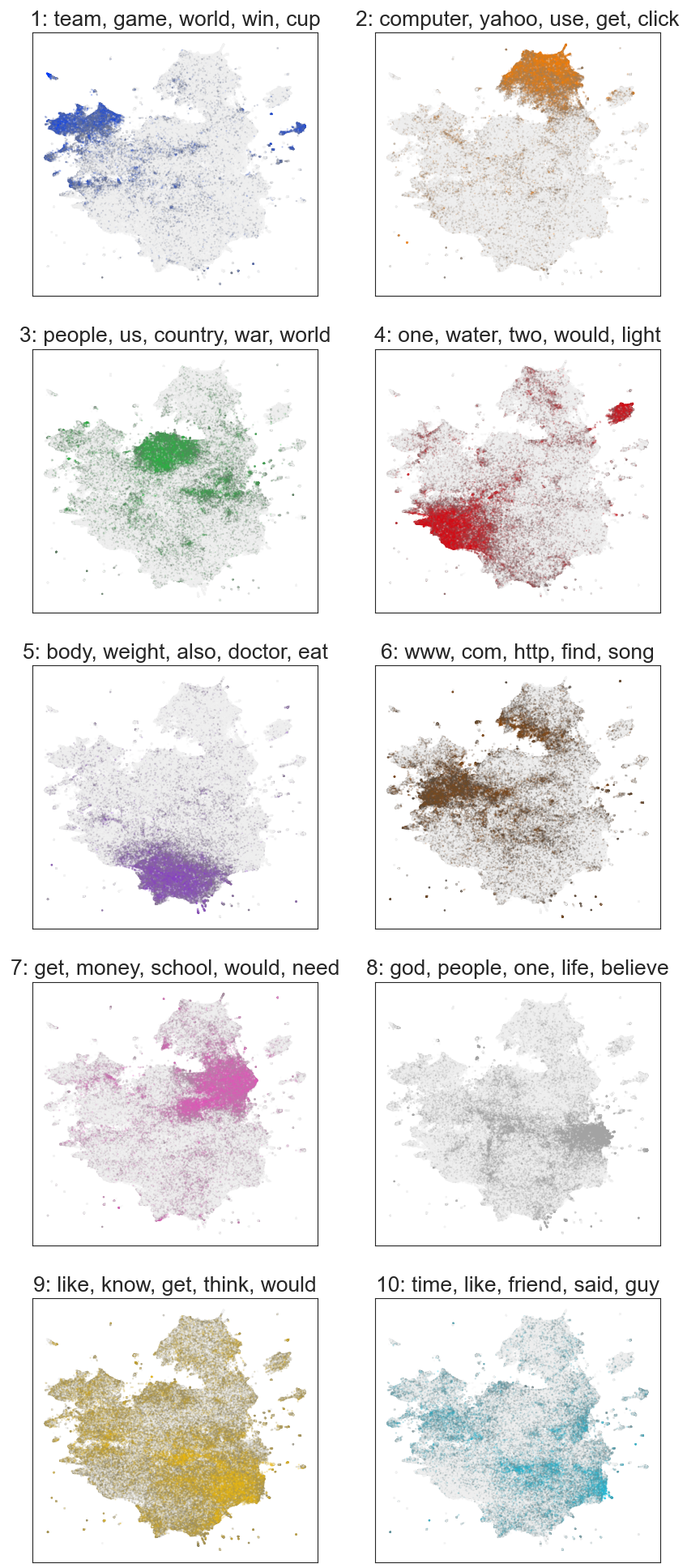}
\caption{Topic proportion of each LDA topic from Table \ref{tab:lda_topics_yahoo} across all \emph{Yahoo Answers} posts in the semantic embedding. The topics are ordered by decreasing information gain. The 300 dimension document vectors are embedded into 2 dimensions using UMAP.}
\label{fig:lda_topics_strengths_yahoo}
\end{figure}

Figure \ref{fig:lda_topics_strengths_yahoo} shows the strengths of each of the 10 LDA topics from Table \ref{tab:lda_topics_yahoo} across all posts. This visually demonstrates that more informative topics are localized in the semantic space, and that LDA discovers topics that are less localized than \texttt{top2vec} topics. Additionally highly unlocalized LDA topics like 9 and 10, which contain the lowest information gain scores, also contain generic words that would not be considered topical or informative by a user. Figure \ref{fig:lda_topics_strengths_yahoo} demonstrates visually that, apart from LDA topics containing less informative words, the reason LDA topics receive lower topic information gain is that they are less localized than \texttt{top2vec} topics and therefore less informative.

\section{Discussion}

We have described \texttt{top2vec}, an unsupervised learning algorithm that finds topic vectors in a semantic space of jointly embedded document and word vectors. We have shown that the semantic space is a continuous representation of topics that allows for the calculation of topic vectors from dense areas of highly similar documents, topic size, and for hierarchical topic reduction. The \texttt{top2vec} model also allows for comparing similarity between words, documents and topics based on distance in the semantic space.

We have proposed a novel method for evaluating topics that uses mutual information to calculate how informative topics are of documents. The topic information gain measures the amount of information gained about the documents when described by their topic words. This measures both the quality of topic words and the assignment of topics to the documents. Our results show that \texttt{top2vec} consistently finds topics that are more informative and representative of the corpus than LDA and PLSA, for varying sizes of topics and number of top topic words.

There are several advantages of \texttt{top2vec} over traditional topic modeling methods like LDA and PLSA. The primary advantages are that it automatically finds the number of topics and finds topics that are more informative and representative of the corpus. As demonstrated, stop-word lists are not required to find informative topic words, making it easy to use on a corpus of any domain or language. The use of distributed representations of words alleviates several challenges of traditional methods that use BOW representations of words, which ignore word semantics. 

Traditional topic modeling techniques like LDA and PLSA are generative models; they seek to find topics that recreate the original documents word distributions with minimal loss. This necessitates these models to place uninformative words in topics with high probability, as they make up a large proportion of all documents. Additionally, there is no guarantee that they will find topics that are representative of the corpus. The results show they can find topics that are extremely specific or overly broad.

In contrast, the words closest to \texttt{top2vec} topic vectors are the words that are most \emph{informative} of the documents the topic vectors are calculated from. This is due to the learning task that generates joint document and word vectors, which predicts the document a word came from. This learning task necessitates document vectors to be placed close to the words that are most informative of the documents. The continuous representation of topics in the semantic space allows topic vectors to be calculated from dense areas of those documents. Thus \texttt{top2vec} topics are the words that are most informative of a document, rather than the set of words that recreate the documents distribution of words with accurate proportions. We suggest that \texttt{top2vec} is more appropriate for finding informative and representative topics of a corpus than probabilistic generative models like LDA and PLSA.

The \texttt{top2vec} code is available as an open-source project\footnote{https://github.com/ddangelov/Top2Vec}.

\newpage

\null
\vfill
\begin{table}[h]
\centering
 \caption{Topic information gain for the top 10 words from \texttt{top2vec} topics trained on the \emph{20 news groups} dataset with 20 topics.}
  \begin{tabular}{lll}
    \toprule
    \textbf{Topic Number} & \textbf{Topic Words}     & \textbf{PWI(T)} \\
    \midrule
    \textbf{1} & pitching, pitchers, pitcher, hitter, batting, hit, hitters, baseball, batters, inning        & 74.2   \\ 
    \addlinespace[0.2cm]
    \textbf{2} & bike, ride, riding, bikes, motorcycle, bikers, helmet, riders, countersteering,     & 71.9   \\ 
    & passenger & \\
    \addlinespace[0.2cm]
    \textbf{3} & circuit, voltage, circuits, resistor, signal, khz, impedance, analog, diode, resistors       & 69.1   \\ 
    \addlinespace[0.2cm]
    \textbf{4} & centris, ram, mhz, quadra, nubus, vram, iisi, lciii, cpu, fpu                     & 62.4   \\ 
    \addlinespace[0.2cm]
    \textbf{5} & patient, symptoms, patients, doctor, disease, treatment, jxp, therapy, skepticism,       & 59.1   \\ 
    & physician & \\
    \addlinespace[0.2cm]
    \textbf{6} & koresh, fbi, compound, batf, davidians, atf, waco, raid, fire, bd            & 54.7 \\
    \addlinespace[0.2cm]
    \textbf{7} & israel, arab, arabs, israeli, jews, palestinians, israelis, war, peace, occupied & 54.3 \\
    \addlinespace[0.2cm]
    \textbf{8} & orbit, space, launch, orbital, satellites, lunar, shuttle, spacecraft, moon, earth & 53.7 \\
    \addlinespace[0.2cm]
    \textbf{9} & clipper, nsa, encryption, encrypted, secure, keys, crypto, algorithm, escrow, scheme & 51.6 \\
    \addlinespace[0.2cm]
    \textbf{10} & controller, drives, drive, ide, scsi, floppy, bios, disk, jumpers, esdi & 50.3 \\
    \addlinespace[0.2cm]
    \textbf{11} & windows, drivers, ati, cica, driver, exe, card, autoexec, mode, ini & 50.1 \\
    \addlinespace[0.2cm]
    \textbf{12} & car, engine, cars, ford, brakes, honda, tires, valve, wheel, rear & 49.7 \\
    \addlinespace[0.2cm]
    \textbf{13} & hockey, playoffs, nhl, game, season, team, playoff, teams, scoring, play & 48.0 \\
    \addlinespace[0.2cm]
    \textbf{14} & gun, guns, firearms, laws, weapons, handgun, crime, amendment, handguns, firearm        & 46.6 \\
    \addlinespace[0.2cm]
    \textbf{15} & window, application, xlib, manager, openwindows, motif, server, xview, client, clients     & 43.6    \\ 
    \addlinespace[0.2cm]
    \textbf{16} & jesus, christ, god, bible, church, scripture, christians, scriptures, christian, heaven        & 38.9    \\ 
    \addlinespace[0.2cm]
    \textbf{17} & postscript, format, printer, fonts, files, formats, font, truetype, bitmap, image           & 36.7    \\
    \addlinespace[0.2cm]
    \textbf{18} & shipping, sale, offer, condition, asking, brand, sell, obo, price, selling & 36.0 \\
    \addlinespace[0.2cm]
    \textbf{19} & atheists, belief, religion, beliefs, god, christianity, truth, religions, believe, atheist & 34.4    \\ 
    \addlinespace[0.2cm]
    \textbf{20} & please, mail, post, email, posting, address, thanks, reply, interested, appreciate                  & 11.3   \\
     & & \textunderscore\textunderscore\textunderscore\textunderscore \\
    \addlinespace[0.2cm]
   & & \textbf{996.6} \\
    \bottomrule
  \end{tabular}
  \label{tab:top2vec_topics}
\end{table}
\vfill

\begin{table}[h]
\centering
 \caption{Topic information gain for the top 10 words from LDA topics, \emph{after} stop-word removal, trained on the \emph{20 news groups} dataset with 20 topics.}
  \begin{tabular}{lll}
    \toprule
    \textbf{Topic Number} & \textbf{Topic Words}     & \textbf{PWI(T)} \\
    \midrule
    \textbf{1} & la, pit, gm, det, bos, tor, pts, chi, vs, min        & 49.9  \\ 
    \addlinespace[0.2cm]
    \textbf{2} & hz, cx, ww, uw, qs, c\_, pl, lk, ck, ah               & 47.0   \\ 
    \addlinespace[0.2cm]
    \textbf{3} & ax, max, pl, di, tm, ei, giz, wm, bhj, ey            & 42.6   \\ 
    \addlinespace[0.2cm]
    \textbf{4} & db, period, goal, play, pp, shots, st, power, mov, bh                     & 27.9   \\ 
    \addlinespace[0.2cm]
    \textbf{5} & mk, mm, mp, mh, mu, mr, mj, mo, mq, mx               & 27.7   \\ 
    \addlinespace[0.2cm]
    \textbf{6} & health, medical, new, study, research, disease, cancer, use, patients, drug & 19.0 \\
    \addlinespace[0.2cm]
    \textbf{7} & armenian, people, said, one, armenians, turkish, went, us, children, turkey & 17.3 \\
    \addlinespace[0.2cm]
    \textbf{8} & dos, windows, drive, card, system, disk, mb, scsi, pc, mac & 16.7 \\
    \addlinespace[0.2cm]
    \textbf{9} & file, program, window, files, image, jpeg, use, windows, display, color & 15.7 \\
    \addlinespace[0.2cm]
    \textbf{10} & government, president, law, would, mr, israel, state, rights, fbi, states & 13.4 \\
    \addlinespace[0.2cm]
    \textbf{11} & god, jesus, bible, church, christian, christ, christians, faith, lord, man & 12.2 \\
    \addlinespace[0.2cm]
    \textbf{12} & game, team, games, hockey, season, teams, league, nhl, new, players & 12.0 \\
    \addlinespace[0.2cm]
    \textbf{13} & space, nasa, earth, launch, shuttle, orbit, moon, satellite, solar, mission & 11.8 \\
    \addlinespace[0.2cm]
    \textbf{14} & edu, ftp, graphics, available, pub, image, mail, com, version, also & 9.7 \\
    \addlinespace[0.2cm]
    \textbf{15} & would, know, anyone, get, thanks, like, one, please, help, could & 8.5 \\
    \addlinespace[0.2cm]
    \textbf{16} & key, use, data, system, one, information, may, encryption, used, number             & 7.5    \\ 
    \addlinespace[0.2cm]
    \textbf{17} & people, would, one, think, know, like, say, even, see, way        & 7.3    \\ 
    \addlinespace[0.2cm]
    \textbf{18} & one, car, would, like, get, time, much, also, back, power            & 7.1    \\ 
    \addlinespace[0.2cm]
    \textbf{19} & edu, com, please, list, mail, sale, send, email, price, offer                  & 4.0   \\ 
    \addlinespace[0.2cm]
    \textbf{20} & think, year, would, good, time, last, well, get, one, got & 3.6    \\ 
    & & \textunderscore\textunderscore\textunderscore \\
    \addlinespace[0.2cm]
    & & \textbf{360.9} \\
    \bottomrule
  \end{tabular}
  \label{tab:lda_topics}
\end{table}

\begin{table}[h]
 \caption{Topic information gain for the top 10 words from \texttt{top2vec} topics trained on the \emph{Yahoo Answers} dataset with 10 topics.}
  \centering
  \begin{tabular}{lll}
    \toprule
    \textbf{Topic Number} & \textbf{Topic Words}     & \textbf{PWI(T)} \\
    \midrule
    \textbf{1} & overwrite, rebooting, debug, debugging, reboot, executable, compiler, winxp, xp, winnt   & 112.2   \\ 
    \addlinespace[0.2cm]
    \textbf{2} & securities, unpaid, equity, purchaser, payment, broker, underwriting, issuer, payable,  & 104.4   \\ 
    & underwriter  & \\
    \addlinespace[0.2cm]
    \textbf{3} & regimen, discomfort, inflammation, swelling, psoriasis, puffiness, inflammatory,    & 98.1   \\ 
    & irritation, edema, hypertension & \\
    \addlinespace[0.2cm]
    \textbf{4} &realtionship, realationship, insecurities, confide, hurtful, inlove, clingy, friendship,   & 92.5   \\ 
    & bestfriend, friendships & \\
    \addlinespace[0.2cm]
    \textbf{5} & song, sings, singer, sang, artist, duet, album, lyrics, ballad, vocalist      & 91.9   \\ 
    \addlinespace[0.2cm]
    \textbf{6} & scripture, believers, righteousness, righteous, pious, spiritual, spirituality, sinful,  &  82.2\\
    & worldly, discernment & \\
    \addlinespace[0.2cm]
    \textbf{7} & team, players, game, teams, scoring, league, teammate, scorers, playoff, defensively & 80.0 \\
    \addlinespace[0.2cm]
    \textbf{8} & courses, subjects, curriculum, students, teaching, faculty, syllabus, academic,   & 78.6 \\
    & undergraduate, baccalaureate & \\
    \addlinespace[0.2cm]
    \textbf{9} & war, leaders, politicians, government, democracy, political, terrorists, terrorism,  & 64.6 \\
    & partisan, policies & \\
    \addlinespace[0.2cm]
    \textbf{10} & thus, constant, hence, surface, resulting, greater, therefore, becomes, occurs, larger  & 33.1 \\
    \addlinespace[0.2cm]
     & & \textunderscore\textunderscore\textunderscore\textunderscore \\
    \addlinespace[0.2cm]
   & & \textbf{837.6} \\
    \bottomrule
  \end{tabular}
  \label{tab:top2vec_topics_yahoo}
\end{table}

\begin{table}[h]
 \caption{Topic information gain for the top 10 words from LDA topics, \emph{after} stop-word removal, trained on the \emph{Yahoo Answers} dataset with 10 topics.}
  \centering
  \begin{tabular}{lll}
    \toprule
    \textbf{Topic Number} & \textbf{Topic Words}     & \textbf{PWI(T)} \\
    \midrule
    \textbf{1} & team, game, world, win, cup, play, de, football, best, player   & 23.2   \\ 
    \addlinespace[0.2cm]
    \textbf{2} & computer, yahoo, use, get, click, internet, free, com, need, windows   & 22.3  \\ 
    \addlinespace[0.2cm]
    \textbf{3} & people, us, country, war, world, would, american, bush, government, america   & 18.0   \\ 
    \addlinespace[0.2cm]
    \textbf{4} & one, water, two, would, light, number, energy, used, earth, use  & 16.9   \\ 
    \addlinespace[0.2cm]
    \textbf{5} & body, weight, also, doctor, eat, blood, may, day, get, pain      & 15.7   \\ 
    \addlinespace[0.2cm]
    \textbf{6} & www, com, http, find, song, name, know, anyone, org, music     & 14.2 \\
    \addlinespace[0.2cm]
    \textbf{7} & get, money, school, would, need, work, pay, good, business, job & 13.1 \\
    \addlinespace[0.2cm]
    \textbf{8} & god, people, one, life, believe, jesus, word, many, would, us  & 12.5 \\
    \addlinespace[0.2cm]
    \textbf{9} & like, know, get, think, would, want, people, good, really, go & 9.1 \\
    \addlinespace[0.2cm]
    \textbf{10} & time, like, friend, said, guy, back, would, one, years, got  & 8.3 \\
    \addlinespace[0.2cm]
     & & \textunderscore\textunderscore\textunderscore\textunderscore \\
    \addlinespace[0.2cm]
   & & \textbf{153.3} \\
    \bottomrule
  \end{tabular}
  \label{tab:lda_topics_yahoo}
\end{table}

\clearpage
\bibliographystyle{unsrt}
\bibliography{references}  


\end{document}